%% file: main.tex
\documentclass[]{article}

\usepackage[margin=1in]{geometry}
\usepackage[utf8]{inputenc} 
\usepackage[T1]{fontenc}    
\usepackage{hyperref}       
\hypersetup{
    colorlinks = true,
    citecolor = blue,
    linkcolor = red
}
\usepackage{url}            
\usepackage{booktabs}       
\usepackage{amsfonts}       
\usepackage{nicefrac}       
\usepackage{microtype}      
\usepackage{xcolor}         

\usepackage{color}
\usepackage{float}
\usepackage[style=base]{caption}
\usepackage{subcaption}
\usepackage{tikz}
\usepackage{tikz-cd}
\usepackage{pgfplots}
\usepackage{abstract}
\usepgfplotslibrary{groupplots}
\pgfplotsset{compat=1.18}

\usepackage{amsmath,amssymb,amsthm,mathtools,bbm}
\usepackage{physics}
\usepackage{dsfont}
\usepackage{mathrsfs}
\usepackage[ruled,vlined,linesnumbered]{algorithm2e}
\usepackage[capitalize]{cleveref}
\usepackage{xfrac}
\usepackage{graphicx}
\usepackage{authblk}
\usepackage{natbib}

\graphicspath{{imgs/}}

\usepackage{comment}

\usepackage{enumitem}
\setlist[enumerate]{leftmargin=1.5em}
\setlist[itemize]{leftmargin=1.5em}
\usepackage{autonum}

\makeatletter
\renewenvironment{proof}[1][]
{%
  \par\noindent{\bfseries\upshape 
    \proofname\if\relax\detokenize{#1}\relax\else\ of #1\fi:\space}%
}
{\qedsymbol}
\makeatother

\input{math-definitions}

\theoremstyle{plain} 
\newtheorem{theorem}{Theorem}[section]
\newtheorem{lemma}[theorem]{Lemma}
\newtheorem{proposition}[theorem]{Proposition}
\newtheorem{corollary}[theorem]{Corollary}

\theoremstyle{definition} 
\newtheorem{definition}{Definition}[section]
\newtheorem{assumption}{Assumption}[section]

\theoremstyle{remark} 
\newtheorem{remark}{Remark}[section]
\newtheorem{example}[remark]{Example}

\crefname{assumption}{Assumption}{Assumptions}

\author[1]{Matteo Vilucchio}
\author[1,2]{Yatin Dandi}
\author[1]{Matéo Pirio Rossignol}
\author[3]{Cedric Gerbelot}
\author[1]{Florent Krzakala}

\affil[1]{Information Learning and Physics Laboratory, \'Ecole Polytechnique F\'ed\'erale de Lausanne (EPFL)}
\affil[2]{Statistical Physics of Computation Laboratory, \'Ecole Polytechnique F\'ed\'erale de Lausanne (EPFL)}
\affil[3]{Unité de Mathématiques Pures et Appliquées, \'Ecole Normale Sup\'erieure de Lyon (ENS)}

\title{Asymptotics of Non-Convex Generalized Linear Models in High-Dimensions: A proof of the replica formula}

\begin{document}
\maketitle

\begin{abstract}
The analytic characterization of the high-dimensional behavior of optimization for Generalized Linear Models (GLMs) with Gaussian data has been a central focus in statistics and probability in recent years. While convex cases, such as the LASSO, ridge regression, and logistic regression, have been extensively studied using a variety of techniques, the non-convex case remains far less understood despite its significance. A non-rigorous statistical physics framework has provided remarkable predictions for the behavior of high-dimensional optimization problems, but rigorously establishing their validity for non-convex problems has remained a fundamental challenge. In this work, we address this challenge by developing a systematic framework that rigorously proves replica-symmetric formulas for non-convex GLMs and precisely determines the conditions under which these formulas are valid. Remarkably, the rigorous replica-symmetric predictions align exactly with the conjectures made by physicists, and the so-called \textit{replicon condition}. The originality of our approach lies in connecting two powerful theoretical tools: the Gaussian Min-Max Theorem, which we use to provide precise lower bounds, and Approximate Message Passing (AMP), which is shown to achieve these bounds algorithmically. We demonstrate the utility of this framework through significant applications: (i) by proving the optimality of the Tukey loss over the more commonly used Huber loss under a $\varepsilon$ contaminated data model, (ii) establishing the optimality of negative regularization in high-dimensional non-convex regression and (iii) characterizing the performance limits of linearized AMP algorithms. By rigorously validating statistical physics predictions in non-convex settings, we aim to open new pathways for analyzing increasingly complex optimization landscapes beyond the convex regime.
\end{abstract}

\section{Introduction}

High-dimensional optimization problems are ubiquitous in modern machine learning and statistics, with applications ranging from sparse regression, to kernels, to neural networks. In statistics, mathematics, and theoretical machine learning, a large body of works have been dedicated to the characterization of the analytical asymptotic performance with synthetic data (see e.g. \citep{donoho2009observed,bayati2011lasso,thrampoulidis2018precise,sur2019modern,thrampoulidis2020theoretical,loureiro2021learning,mei2022generalization,tan2024multinomial}). While substantial progress has been made in understanding convex problems, such as LASSO \citep{Donoho_AMP_compress_sensing_2010,bayati2011lasso, miolane2021distribution} or random feature regression \citep{mei_generalization_2022,hu2024asymptotics}, the non-convex landscape remains largely unexplored despite its practical significance. 

We shall be interested here in the \textit{asymptotic performances} of \textit{non convex} generalized linear models in high-dimension.  We consider a supervised learning task where we are given $n$ input output pairs in the form of a dataset $\dataset = \{(\x_\dataidx,y_\dataidx)\}_{\dataidx \in [n]}$ where each one of the pairs $(\x_\dataidx,y_\dataidx) \in \RR^{d} \times \RR$ and we perform an estimation of parameters $\what \in \mathbb{R}^{d}$ through the minimization of a (possibly non-convex) objective. We define the minimization objective 
\begin{equation}\label{eq:risk-def}
    \riskfun_d(\w) \defeq \frac{1}{d} \qty[ 
        \sum_{\dataidx=1}^n 
        \lossfun\qty(y_\dataidx, \frac{\w^\top \x_\dataidx}{\sqrt{d}} ) + 
        \sum_{\dimidx=1}^{d} \regfun(\w_\dimidx) 
    ] \,,
\end{equation}
where $\lossfun : \RR \cross \RR \to \RR$ is the loss function, $\regfun : \RR \to \RR$ a regularization function.  Given hyperparameters $0 \leq a \leq b \leq \infty$, we study the constrained optimization problem
\begin{align}
    \minriskfun_d &\defeq \min_{\w \in \mathcal{K}_{a,b}}  \riskfun_d(\w) \,, \label{eq:minimum-risk} \\
    \wminimizer &\in \argmin_{\w \in \mathcal{K}_{a,b}}  \riskfun_d(\w) \,, \label{eq:minimiser-risk}
\end{align}
where $\mathcal{K}_{a,b} \defeq \{\w \in \RR^d : a \leq \frac{1}{d}\|\w\|_2^2 \leq b\}$ defines a spherical shell constraint. When $b = \infty$, the upper bound constraint is omitted, resulting in $\mathcal{K}_{a,\infty} = \{\w \in \RR^d : a \leq \frac{1}{d}\|\w\|_2^2\}$. Note that for $a > 0$, the constraint set $\mathcal{K}_{a,b}$ is non-convex, introducing an additional source of non-convexity beyond the potential non-convexity of $\lossfun$ and $\regfun$.

We consider that the input-output pairs that make up the dataset are sampled independently from a distribution $P_{\wteacher}(\x, y)$. 
This probability distribution depends on a target linear model  parameterized by $\wteacher \in \RR^d$ called teacher weights. Specifically the \emph{generative} setting is the one where
\begin{align}\label{eq:data-distribution}
    P(\x, y) = \pout{
        y \mid \frac{\langle \wteacher, \x \rangle
    }{\sqrt{d}}}P_{\mathrm{in}}(\x) \,,
\end{align}
where $P_{\mathrm{in}}$ is a probability density function over $\RR^{d}$ and $\pout{ \cdot \mid z} : \RR \to \RR$ encodes our assumption that the label is a (potentially non-deterministic) function of $\langle \wteacher, \x \rangle / \sqrt{d}$. 
We equivalently say that the output labels are generated by $y_\dataidx = \outputfun(\frac{1}{\sqrt{d}}\langle \wteacher, \x_\dataidx \rangle)$, where $\outputfun : \mathbb{R} \to \mathbb{R}$ is possibly stochastic.
In the rest of the manuscript we will consider the Gaussian case where the samples $\x$ have Gaussian distribution $P_\mathrm{in} = \mathcal{N}(\mathbf{0}, \operatorname{Id}_d)$. 

The convex setting is well understood, as powerful theoretical tools have emerged to analyze these high-dimensional problems. Two such important tools, in particular, are the Gaussian min-max theorem \citep{gordon_1988,stojnic2009various,thrampoulidis2015gaussian,loureiro2021learning}, and the analysis of fixed-points of the approximate message passing algorithms (AMP) \citep{Donoho_AMP_compress_sensing_2010,bayati2011lasso,bolthausen2014iterative,Loureiro_gaussian_mixtures_2021}. Another, different, approach, has been through analyses based on random matrix theory \citep{dobriban2018high,richards_2021a,bach2024high}. However, extending these results beyond convex setting has remained a fundamental challenge, as the existing techniques heavily rely on convex analysis.

Yet, there are strong indication that these limits are not fundamental. Indeed powerful heuristic have been developed for these problem by the statistical physics community though the use of the replica and cavity methods \citep{mezard1987spin,mezard2009information,advani2013statistical,zdeborova2016statistical}. These methods predict a whole range of formula \textit{very similar} to the ones appearing in convex cases,  but it is notably difficult to establish their validity \citep{talagrand2003spin}. In particular, convexity does not play a fundamental role in these predictions, which instead rely on the so-called \textit{replica symmetric} assumption. 

In this work, we bridge this gap, and provide \textit{a rigorous proof} of the statistical physics predictions for the loss and the generalization predicted by these formulae in the non-convex case, that matches exactly the validity range of the replica symmetric assumption predicted by physicists. The originality of our approach is that we use a combination of both the  Gaussian min-max theorem, and the message passing techniques. Specifically we use Gaussian min-max theorem as a \textit{lower bound for the true minimum of the optimization problem}. This true minimum can be characterized as the solution of a low dimensional problem. Later, \textit{we show how this bound can be achieved through a carefully designed AMP algorithm} that depends only on the solution of the low dimensional problem, and whose convergence can be checked from a low dimensional integral depending on the solution of the low dimensional problem. This framework not only provides rigorous mathematical guarantees but also validates numerous predictions from statistical physics that have long suggested the existence of such characterizations beyond the convex regime. 

These low-dimensional fixed-point equations we provide emerge naturally from both the Gaussian min-max theorem analysis and the state evolution of AMP algorithms. 
Furthermore, our analysis reveals that the validity of these characterizations does not  depend on convexity. Instead, it relies on a more subtle stability condition (known in physics as the \textit{replicon} or \textit{de Almeida-Thouless} condition \citep{de1978stability,de1983eigenvalues}) that can be verified directly --- and trivially --- from the low-dimensional equations. Interestingly, this condition turns out to be the same as the point-wise stability convergence criterion of the message passing algorithm \citep{bolthausen2014iterative}. The replicon condition, which we fully characterize, determines when the high-dimensional optimization landscape is sufficiently well-behaved for our analysis to hold. Notably, this condition can be satisfied even in clearly non-convex settings, explaining the empirically observed success of certain optimization procedures in non-convex high-dimensional problems. 
The validity of our results are precisely those conjectured to be necessary by statistical physics: when this condition is not satisfied
the replica symmetric formula are known to be wrong (and needs to be replaced by the more involved replica symmetry broken ones \citep{talagrand2006parisi}, which are beyond the scope of this paper).

We demonstrate the power of this approach through some applications that have attracted significant recent attention: negative regularization in high-dimensional regression for generic losses \citep{tsigler2023benign,kobak2020optimal}, the performance of spectral algorithms
\citep{mondelli2018fundamental,lu2020phase,maillard2022construction}, and the study of non convex losses \citep{loh2017statistical,chang2018robust} in robust training \citep{vilucchio2024asymptotic,adomaityte2024high}. Our results open new pathways for analyzing increasingly complex optimization landscapes in high-dimensional statistics, offering both theoretical insights and practical algorithmic implications for modern machine learning problems.

\subsection{Further related works}
Our work synthesizes and extends several fundamental theoretical frameworks in high-dimensional statistics and optimization, as well as in the mathematical spin glass litterature.
\paragraph{The Gaussian Min-Max Theorem (GMT) ---} The GMT \citep{gordon_1988}, later strengthened to the Convex GMT \citep{thrampoulidis2015gaussian}, is a powerful tool for analyzing high-dimensional problems. 
While it has proven highly effective for convex problems \cite{thrampoulidis2018precise,javanmard_precise_statistical_2022,loureiro2021learning}, its application to non-convex settings has remained limited. 
Our work demonstrates how GMT can be extended beyond convexity through careful analysis of the associated low-dimensional optimization landscape.

\paragraph{Approximate Message Passing (AMP) ---} Originating in spin glass theory as the Thouless-Anderson-Palmer (TAP) equations \citep{thouless1977solution} for the Sherrington-Kirpatrick model \citep{mezard1987spin}, these message passing algorithms have witnessed a burst of popularity after the work of \citet{bolthausen2014iterative} and its extension to compressed sensing in \citet{Donoho_AMP_compress_sensing_2010}. They have become a cornerstone of high-dimensional inference with synthetic data. The precise analysis of AMP dynamics \citep{bayati2011lasso} revealed that these algorithms can often achieve fundamental statistical limits. Recent work has extended the analysis of AMP to a broader class of inference problems through oriented graph representations \citep{gerbelot2023graph}.  
Related to us are the works of \citet{montanari2015non}, who characterized phase transitions in estimation error for non-negative PCA, and \citet{celentano_2023_z2,celentano2024sudakov}, who established local convexity of the so-called TAP free energy and finite-sample convergence guarantees for Z2-synchronization.
Their approach also relies on a combination of Gaussian comparison inequalities and message passing iterations, albeit for a restricted class of problems.
In our approach, we use AMP across a much broader class of non-convex problems.

\paragraph{Spin glass theory ---}
Our work builds upon fundamental results in spin glass theory. The mathematical treatment of replica symmetry was pioneered by \citet{talagrand2003spin}. 
This was later extended to the full Parisi solution \citep{talagrand2006parisi,panchenko2013sherrington}, establishing the exactness of the replica method predictions. In this work, we focus on the simpler replica symmetric regime.

\paragraph{Applications ---} Our framework also yields new insights in two important areas that have attracted significant recent attention. 
The first concerns negative regularization, which has been primarily studied in the specific context of ridge regression \citep{hastie_2022_surprises,NEURIPS2020_72e6d323,richards_2021a}, and the analysis can be carried out because of the quadratic loss. 
Our results extend these insights to general loss functions, providing an interest in robust statistics where non-convex losses such as Tukey's biweight and Cauchy loss functions are used for handling outliers \citep{lozano2013minimum,loh2024theoretical}.
Another application provides a comprehensive analysis of linearized AMP algorithms, and the performance of spectral method, where we show we can recover directly the results from \citep{montanari2015non,mondelli2018fundamental,lu2020phase,maillard2022construction}, without resorting to random matrix theory tools. 

\paragraph{Notation:} Given a function $f : \RR \to \RR$, we define its Moreau envelope $\mathcal{M}_{f} : \RR \to \RR$ as 
\begin{equation}\label{eq:main:moreau-definition}
    \mathcal{M}_{f}(\omega) = \inf_{z \in \RR} \qty[\frac{1}{2} (\omega - z)^2 + f(z)] \,.
\end{equation}
The associated proximal operator $\mathcal{P}_{f} : \RR \to \mathscr{P}(\RR)$ is the set-valued mapping defined as
\begin{equation}\label{eq:main:proximal-definition}
    \mathcal{P}_{f}(\omega) = \argmin_{z \in \RR} \qty[\frac{1}{2} (\omega - z)^2 + f(z)] \,,
\end{equation}
where $\mathscr{P}(\RR)$ denotes the power set of $\RR$. When the minimizer is unique, $\mathcal{P}_{f}(\omega)$ is singleton-valued and more specifically when the function $f$ is convex we recover the usual proximal function.

We define a function $f:\RR \to \RR$ to be \emph{coercive} if $\lim_{\abs{x} \to \infty} f(x) = \infty$ and the function $f$ is said to be \emph{lower semi-continuous} if $\liminf_{x \to x_0} f(x) \geq f(x_0)$ for all values $x_0$ where it is defined.
Additionally we say that a function $f: \RR \to \RR$ is \emph{$V_0$-weakly convex} for $V_0 \in [0, \infty)$ if the function $x \mapsto V_0 f(x) + \frac{x^2}{2}$ is convex. 
Additionally we say that a function $f$ is said to be \emph{prox-coercive} if for any $\delta \in \RR$, the function $x \mapsto f(x) + \frac{(\delta - x)^2}{2}$ is coercive in $x$.

\section{Statement of Main Result}\label{sec:statement-main-result}

Our main result characterizes the asymptotic behavior of the minimum and minimizers of the high-dimensional optimization problem~\cref{eq:risk-def} through a low-dimensional auxiliary problem involving six order parameters. Before stating this result precisely, we establish our model assumptions and introduce the necessary notation.
Our goal is to study the behavior in the 
asymptotic regime

\begin{assumption}[High-dimensional limit]\label{ass:high-dimensional-limit}
We consider the limit where $n,d \to \infty$ and their ratio is kept fixed at $n / d \to \alpha \in (0,\infty)$.
We will refer to $\alpha$ as the \textit{sample complexity}.
\end{assumption}

We also make the following assumption on the channel generating the data

\begin{assumption}[Teacher second moment and Output channel]\label{ass:teacher-channel}
For each $d$, let $\wteacher \in\RR^d$ be a teacher vector satisfying $\|\wteacher\|_2^2 / d \to \rho \in (0,\infty)$ almost surely as $d\to\infty$.
Moreover, the empirical distribution of the coordinates of $\wteacher$ converges almost surely to a probability law $P_{\tilde w}$ on $\RR$ with $\mathbb E[\tilde w^2]=\rho$.

Let $\x\sim\mathcal N(0,\operatorname{Id}_d)$ and define $s\defeq \langle \wteacher,\x\rangle/\sqrt d$.
Conditionally on $s$, the response $y$ is generated according to an output channel
\begin{equation}
    y \sim P_{\mathrm{out}}(\cdot \mid s) \,,
\end{equation}
where $P_{\mathrm{out}}$ is fixed (independent of $d$) and such that $y$ is almost surely bounded.
\end{assumption}

We consider loss and regularization functions satisfying the following assumption
\begin{assumption}[Regularity of loss and regularization functions]\label{ass:lossregfun}
    The functions $\lossfun$ and $\regfun$ are pseudo-Lipschitz of finite order (see \Cref{def:pseudo-lipshitz-fucntion}) and respectively $V_L$ and $V_R$-weakly convex (see \Cref{def:weakly-convex-definition}) for some $V_L, V_R > 0$.
    When $b = \infty$, we further require that both $\lossfun$ and $\regfun$ are bounded below
    and that there exist constants $C_R, W_0 > 0$ such that $\regfun(w) \geq C_R w^2$ for all $|w| \geq W_0$.
\end{assumption}

The pseudo-Lipschitz condition allows for non-smooth losses while still maintaining sufficient regularity for our theoretical analysis. When the norm constraint $b$ is infinite, the coercivity condition guarantees the existence of minimizers by preventing the objective from having escaping sequences. 

With these modeling assumptions in place, we now introduce the low-dimensional characterization. We thus define the low-dimensional objective of interest as
\begin{equation}\label{eq:low-dimensional-energy}
    \energyfun(m, q, \tau, \kappa, \nu, \eta) \defeq 
    \frac{\kappa \tau}{2}
    - \frac{1}{2 \eta} (\nu^2 \rho + \kappa^2 )
    + \nu m
    - \frac{\eta q}{2} 
    - \alpha \mathcal{E}_{\lossfun} 
    - \mathcal{E}_{\regfun} \,,
\end{equation}
where
\begin{align}
    \energyfun_\lossfun &\defeq \mathbb{E}_{g,s}\qty[
        \frac{\kappa}{\tau}\mathcal{M}_{\frac{\tau}{\kappa} \lossfun(y, \cdot)}\qty(
            \frac{m}{\sqrt{\rho}} s + \sqrt{q - \frac{m^2}{\rho}} h
        ) 
    ] \,, \label{eq:def-loss-mean-moreau} \\
    \energyfun_\regfun &\defeq \mathbb{E}_{\tilde{w}, h}\qty[
        \eta\mathcal{M}_{\frac{1}{\eta} \regfun(\cdot)}\qty(
            \frac{1}{\eta} \qty(\nu \tilde{w} + \kappa g)
        ) 
    ] \,, \label{eq:def-reg-mean-moreau}
\end{align}
and the expectations are taken with respect to independent standard Gaussian random variables $s,g,h \sim \mathcal{N}(0,1)$, the teacher coordinates $\tilde{w} \sim P_{\tilde w}$, and the output $y \sim P_{\mathrm{out}}(\cdot \mid s)$. The Moreau envelope $\mathcal{M}_{f(\cdot)}$ is defined in~\cref{eq:main:moreau-definition}.

We define $\mathcal{S}^\star$ as the set of extremizers of the following min-max problem:
\begin{equation}\label{eq:def-s-star}
    \mathcal{S}^\star = 
    \qty{ 
        \hat{m}, \hat{q}, \hat{\tau}, \hat{\kappa}, \hat{\nu}, \hat{\eta}: \energyfun(\hat{m}, \hat{q}, \hat{\tau}, \hat{\kappa}, \hat{\nu}, \hat{\eta})=
        \sup_{\kappa, \nu, \eta} 
        \inf_{
            \substack{m, q, \tau \\ a \leq q \leq b}
        }
        \energyfun(m, q, \tau, \kappa, \nu, \eta) 
    } \,.
\end{equation}
Under the assumptions, $\mathcal{S}^\star$ is non-empty and fully characterizes the high-dimensional optimization landscape.

While the previous assumptions can be considered standard, our analysis relies on two additional key assumptions. 
The first ensures regularity of the proximal operators, while the second provides a crucial stability condition that emerges from the statistical physics of complex systems.

\begin{assumption}[Weak Convexity of scaled Loss and Regularization Functions]\label{ass:weak_convexity}
There exist $(m^\star, q^\star, \tau^\star, \kappa^\star, \nu^\star, \eta^\star) \in \mathcal{S}^\star$ such that $\frac{\tau^\star}{\kappa^\star}<V_L$ and $\frac{1}{\eta^\star} < V_R $, i.e. $\frac{\tau^\star}{\kappa^\star}\lossfun(y,\cdot)$ and $\frac{1}{\eta^\star}\regfun $ are weakly convex with modulus $\frac{\kappa^\star}{\tau^\star}V_L >1$ and $\eta^\star V_R >1$ respectively.
\end{assumption}

\begin{remark}
As established in~\Cref{sec:app:from-weak-convex-to-lipshitz}, this assumption is both necessary and sufficient for the existence of Lipschitz continuous proximal selections. 
Specifically, under these conditions, the proximal operators $\mathcal{P}_{\frac{\tau^\star}{\kappa^\star}\lossfun(y,\cdot)}$ and $\mathcal{P}_{\frac{1}{\eta^\star}\regfun}$ admit unique $L_L$-Lipschitz and $L_R$-Lipschitz selections 
\begin{equation}
    \tilde{\mathcal{P}}_{\frac{\tau^\star}{\kappa^\star}\lossfun(y,\cdot)}(\omega) \in \mathcal{P}_{\frac{\tau^\star}{\kappa^\star}\lossfun(y,\cdot)}(\omega), \quad
\tilde{\mathcal{P}}_{\frac{1}{\eta^\star}\regfun}(w) \in \mathcal{P}_{\frac{1}{\eta^\star}\regfun}(w), \quad \forall \omega, w, y \in \RR\,,
\end{equation}
with $L_L\defeq \frac{\kappa^*V_L}{\kappa^*V_L-\tau^*} >1$ and $L_R\defeq \frac{\eta^*V_R}{\eta^*V_R-1} >1$.

This equivalence justifies our choice of assumption as the minimal regularity condition needed for our algorithmic framework.
Similar characterizations have appeared in the literature \citep{gribonval2020characterization,gribonval2021bayesian}, but our result establishes the tightness of this condition.
An example of loss function violating the condition is given by $\lossfun(z) = \mathbf{1}_{z \geq \kappa}$ for some $\kappa \in \mathbb{R}$, corresponding to the setting of perceptron \citep{gardner1988space}.
\end{remark}

The previous assumption and remark allows us to define two Lipschitz-continuous update functions that will be used in the message passing algorithm
\begin{equation}\label{eq:output-input-denoisers-amp}
    \outden[\star](\omega, y) = \frac{\kappa^\star}{\tau^\star} \qty(\tilde{\mathcal{P}}_{\frac{\tau^\star}{\kappa^\star}\lossfun(y,\cdot)}(\omega) -\omega ) \,, \quad
    \inden[\star](w) = \tilde{\mathcal{P}}_{\frac{1}{\eta^\star}\regfun}(w) \,.
\end{equation}

We now introduce the concept of Replicon Stability for the family of problems we consider, as follows:

\begin{definition}[Replicon Stability]\label{def:replicon-stability}
We say that $(m^\star, q^\star, \tau^\star, \kappa^\star, \nu^\star, \eta^\star) \in \mathcal{S}^\star$ already satisfying~\Cref{ass:weak_convexity} are stable if 
\begin{equation}\label{eq:stab}
    \alpha \mathbb{E}_{\tilde{w},z}
    \qty[
        \qty(\partial \inden[\star] \qty(
            \frac{\nu^\star}{\eta^\star} \tilde{w} + \frac{\kappa^\star}{\eta^\star} z) 
        )^2
    ] 
    \mathbb{E}_{y, z}\qty[
        \qty(\partial \outden[\star] \qty( 
            \frac{m^\star}{\rho} s + \sqrt{q^\star - \frac{(m^\star)^2}{\rho}} z, y) 
        )^2
    ] 
    < 1
    \,,
\end{equation}
where $z,s  \sim \mathcal{N}(0,1)$, $y \sim \pout{\cdot \mid s}$, $\tilde{w}$ as in~\Cref{ass:teacher-channel} and $\outden[\star], \inden[\star]$ are defined in~\cref{eq:output-input-denoisers-amp}. 
\end{definition}

Our second assumption ensures stability of the optimization algorithm as follows
\begin{assumption}\label{ass:stab}
We suppose that for some $(m^\star, q^\star, \tau^\star, \kappa^\star, \nu^\star, \eta^\star) \in \mathcal{S}^\star$ the condition in~\Cref{def:replicon-stability} is valid.
\end{assumption}

The above condition corresponds to the replicon condition from statistical physics \citep{EngelVandenBroeck2001} and generalizes the stability criterion of \citet[Eq.~(2.1)]{bolthausen2014iterative}. Most importantly, it provides a necessary condition for the replica symmetric ansatz to remain valid for the setting considered here, making it an essential requirement for our analysis. 
However, the condition itself is not sufficient. 
As an example, for the case of spherical perceptron, corresponding to $\lossfun(z) = \mathbf{1}_{z \geq \kappa}$, the minimum value of the objective deviates from the replica-symmetric prediction even under the validity of the Replicon Stability condition \citep{franz2016simplest,sclocchi2022optimiz}.

Hence, both the \Cref{ass:weak_convexity,ass:stab} are fundamental rather than technical. 

Our main theorem establishes the asymptotic behavior of the high-dimensional optimization problem as follows

\begin{theorem}\label{thm:main}
Consider the sequence of random optimization objectives $\riskfun_d$ defined in \cref{eq:risk-def}.
Under~\Cref{ass:weak_convexity,ass:stab,ass:lossregfun,ass:high-dimensional-limit,ass:teacher-channel}, we have
\begin{enumerate}
    \item%
    {The minimum risk $\minriskfun_d$ concentrates almost surely to a low-dimensional sup-inf problem as 
    \begin{equation}\label{eq:min-risk-limit-def}
        \minriskfun_d 
        \xrightarrow[]{a.s.} 
        \minriskfun \defeq 
        \sup_{\kappa, \nu, \eta} 
        \inf_{
            \substack{m, q, \tau \\ a \leq q \leq b}
        }
        \energyfun(m, q, \tau, \kappa, \nu, \eta) \,.
    \end{equation} 
    }
    \item%
    {For any pseudo-lipschitz $\psi: \mathbb{R}^2 \rightarrow \mathbb{R}$ such that 
    \begin{equation}\label{eq:def_f}
        \Psi(s) \defeq 
        \sup_{\kappa, \nu, \eta} 
        \inf_{
            \substack{m, q, \tau \\ a \leq q \leq b}
        }
        \qty[ 
            \energyfun(m, q, \tau, \kappa, \nu, \eta) + s \psi(m, q) 
        ] 
    \end{equation}
    is differentiable at $s = 0$, any sequence of global minimizers $\what$ satisfies
    \begin{equation}\label{eq:def_psi}
        \psi\qty(\frac{1}{d} \langle \wminimizer, \wteacher \rangle, \frac{1}{d} \norm{\wminimizer}_2^2 ) \xrightarrow[d \to \infty]{a.s.} \eval{\dv{s} \Psi(s) }_{s=0} \,.
    \end{equation}
    }
\end{enumerate}
\ 
\end{theorem}

Our proof combines two complementary approaches. We first establish a lower bound using Gordon's Min-Max Theorem \citep{gordon_1988,thrampoulidis2015gaussian}. We then construct a matching upper bound through an algorithmic realization using a message passing algorithm. 
The main steps of the proof are presented in the following section, while a full derivation is deferred in the appendices. 

The previous theorem presents both an asymptotic characterization of the minimum (first case) and a characterization of function of minimizers (second case) through the choice of the function $\psi$.
A key consequence of this theorem is that it precisely characterizes the high-dimensional estimation error. Indeed, by choosing $\psi: (m, q) \mapsto \rho - 2m + q$, we obtain
\begin{equation}
    \frac{1}{d} \norm{\wminimizer - \wteacher}^2_2 = \psi\qty(\frac{1}{d} \langle \wminimizer, \wteacher \rangle, \frac{1}{d} \norm{\wminimizer}_2^2 ) \,.
\end{equation}

\section{Applications of the Main Result}\label{sec:applications}

We now demonstrate three fundamental settings where our theoretical framework provides novel insights either by exploring a setting that previously was inaccessible for previous tools, proving already known results in the literature or by proving optimality of a loss function in the case of $\varepsilon$ contaminated outlier model.

\begin{figure}[t]
    \centering
    \includegraphics[width=0.45\linewidth]{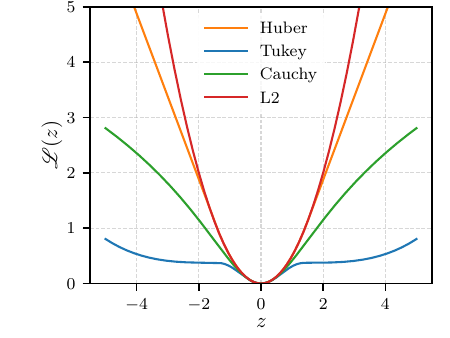}
    \hfill
    \includegraphics[width=0.45\linewidth]{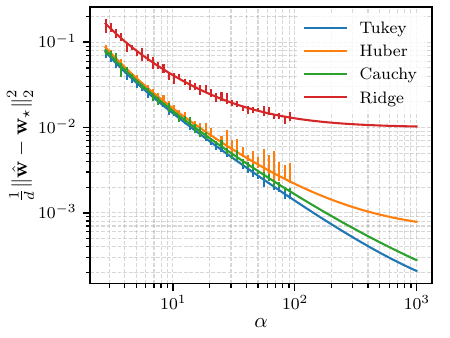}
    \caption{
        \textit{(Left)}
        Presentation of some loss functions used in robust statistics.
        \textit{(Right)}
        Comparison of different performances under a Gaussian $\varepsilon$-contaminated model. We see that non-convex losses attain a lower estimation error than the convex ones. 
        The lines represent the solution for $\psi(m,q) = \rho - 2 m + q$ from~\cref{eq:def_f,eq:def_psi}, while the error bars the ERM simulations on generated data.
    }
    \label{fig:tukey-huber-comparison}
\end{figure}

\subsection{Non Convex Loss Function for Robust M-Estimation}\label{sec:non-convex-losses}

A fundamental challenge in high-dimensional statistics is developing estimators that are robust to outliers while maintaining statistical efficiency. Classical robust statistics has extensively studied this problem in the low-dimensional regime \citep{huber1981robust,maronna2019robust}, with particular focus on M-estimators. 

The design and analysis of loss functions that effectively handle outliers in regression has been central to robust statistics. The classical literature has introduced numerous such functions, emphasizing non-convex alternatives to least squares estimation. Notable examples include Tukey's biweight function \citep{tukey1960survey} and the Cauchy loss \citep{lehmann2006theory}, which offer superior robustness properties compared to convex alternatives. Recent work by \citet{lozano2013minimum} and \citet{loh2017statistical} has renewed interest in this area, studying stationary points of nonconvex functions for robust linear regression within a local neighborhood of the true regression vector.

The theoretical analysis of these estimators in the proportional high-dimensional regime has remained challenging due to their non-convex nature. Our theorem provides a framework to address this challenge, yielding sharp asymptotic characterizations of both convex and non-convex robust regression estimators.

To illustrate our results, we consider the performance of the ERM estimator from~\cref{eq:minimiser-risk} with L2 loss $\lossfun_{\mathrm{L2}}(u) = \frac{u^2}{2}$, the Huber Loss function $\lossfun^\xi_{\mathrm{H}}(u) = \xi^2 \int_0^{\abs{u}/\xi} \min (1,x) \dd{x}$, the Tukey biweight loss function $\lossfun^\xi_{\mathrm{T}}(u)$, and the Cauchy loss function $\lossfun^\xi_{\mathrm{C}}(u)$ defined as
\begin{equation}
    \lossfun^\xi_{\mathrm{T}}(u) = 
    \begin{cases}
        \frac{\xi^2}{6} \qty[1 - \left(1 - \frac{u^2}{\xi^2}\right)^3], & \text{if } |u| \le \xi, \\
        \frac{\xi^2}{6} + \varrho \abs{u - \xi}^3 , & \text{if } u > \xi, \\
        \frac{\xi^2}{6} + \varrho \abs{u + \xi}^3 , & \text{if } u < -\xi.
    \end{cases}
    \,, \quad  
    \lossfun^\xi_{\mathrm{C}}(u) = \frac{\xi^2}{2} \log\left(1 + \frac{u^2}{\xi^2}\right) 
    \,,
\end{equation}
where $u \defeq y - z$ and with $\ell_2$ regularization $\regfun_\lambda(w) = \frac{\lambda}{2} w^2$.
The Tukey loss includes an additional coercive term with parameter $\varrho$, required for our theoretical framework to apply. In practice, for sufficiently small $\varrho$, this modification has negligible effect on the estimator's behavior.

In~\Cref{fig:tukey-huber-comparison} we present a comparison of the optimally tuned regularization parameters for the four different loss functions, showing that for the chosen noise model, the Tukey loss achieves lower estimation error compared to the Huber, Cauchy, and L2 losses.
Since our theory shows that the estimation error converges to a deterministic function of the model parameters, optimizing the hyperparameters $(\xi, \lambda)$ reduces to a two-dimensional optimization problem. While a rigorous justification of this hyperparameter optimization, which requires strengthening our statement from pointwise to uniform convergence, could likely be obtained using techniques similar to those in \citet{miolane2021distribution}, we defer this analysis to a longer version of this work.
Nonetheless, we observe strong agreement between our asymptotic predictions and finite-dimensional simulations.

\begin{figure}[t]
    \centering
    \includegraphics[width=0.48\textwidth]{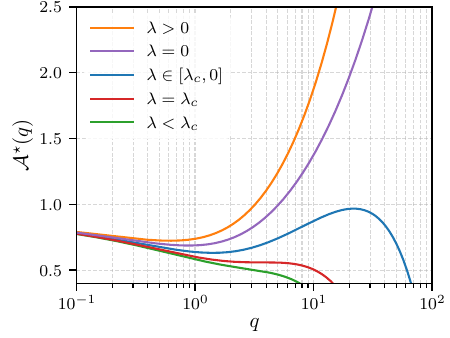}
    \hfill
    \includegraphics[width=0.48\textwidth]{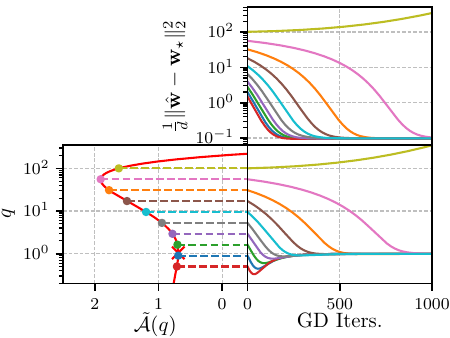}
    \caption{
        \textit{(Left)} 
        Training loss profiles as a function of parameter norm for varying regularization strengths $\lambda$. 
        \textit{(Right)} 
        Gradient descent trajectories in dimension $d=1000$ for different initialization norms. 
        Trajectories initialized with norms smaller than the local maxima converge to the local minimum, while the others diverge.
    }
    \label{fig:negative-regularisation}
\end{figure}

\subsection{Phase Transitions in Negative Regularization for Generalized Linear Models}\label{sec:negative-reg}

We will now consider the phenomenon of negative regularization in high-dimensional GLMs. While this effect has been extensively studied for ridge regression \citep{hastie_2022_surprises,NEURIPS2020_72e6d323,richards_2021a}, its behavior for non-quadratic loss functions and $\ell_2$ regularization remains poorly understood. 
This is a technical hurdle as the previous results strongly rely on Random Matrix Theory.
We provide a characterization of this phenomenon for a broad class of loss functions. 

For a convex $\lossfun(y,z)$ with sub-quadratic growth at infinity, setting $\lambda < 0$ leads the minimization objective to diverge to $-\infty$ as $\norm{\w} \rightarrow \infty$. However, up to a threshold, the objective admits a local minimizer. 
While~\Cref{thm:main} does not directly hold for the case of negative regularization we characterize the training landscape by studying the problem defined in~\cref{eq:risk-def} with $a=b=q$, as a function of $q$.

For a non-negative regularization strength ($\lambda \geq 0$), convexity of the loss ensures the existence of a unique minimizer whose normalized squared norm converges almost surely to a finite value. 
As we move into negative regularization, there exists a critical threshold $\lambda_c < 0$ that separates the case with and without a local minima. 
For $\lambda \in (\lambda_c, 0]$, the loss landscape admits at least two distinct minimizers: a local minimizer at finite norm and a global minimizer whose norm diverges to infinity. 
Below the critical threshold ($\lambda < \lambda_c$), the landscape no longer presents any finite-norm minimizer, and the objective function is minimized through a sequence of solutions with diverging norms.

These theoretical predictions are validated by numerical experiments shown in~\Cref{fig:negative-regularisation} \textit{(Left)} for the Huber loss $\mathscr{L}_{H}^\xi$, which presents the limiting training objective $\minriskfun$ as a function of the norm of the solution $q$ for different values of  $\lambda$. 
For $\lambda > 0$ the loss objective presents a single minima because of convexity.
For $\lambda < 0$ the training loss develops both a local minimum at finite $q$ (which continuously extends from the global minimum at positive $\lambda$) and a global minimum at diverging $q$. 
The minima exists only for values of $\lambda < 0$ not too negative. 
Crucially, these two minima are separated by a local maximum, suggesting that local optimization algorithms initialized at small norm will converge to the non-trivial local minimum. 
Additionally, in~\Cref{fig:negative-regularisation} \textit{(Right)} we consider gradient descent initialized at different norms in the training profile for $\lambda \in (\lambda_c, 0)$. We see that the dynamics of GD follow the landscape already at finite dimension.

\subsection{Spectral Method in Generalized Linear Models}
Last but not least we now present interesting results in the analysis of spectral methods. A standard question is how many observations are required to obtain a better-than-random prediction within a class of algorithms, also known as \emph{weak learnability}. This can be studied both \emph{statistically} (within the class of all, including exponential time algorithms) or \emph{computationally} (restricted to a particular computational class, such as first-order algorithms). For generalized linear models, often called single-index model in this context, weak learnability has been heavily studied. The statistical threshold for learnability has been characterised by 
\cite{barbier_optimal_2019} 
when the covariate dimension $d$ is large. Optimal computational thresholds for the class of first-order iterative algorithms were also derived in \citep{mondelli2018fundamental, luo2019optimal, celentano20a, Maillard2020}. A natural question is wether spectral methods can capture this optimal threshold, and several works have considered this problem, in particular, \citet{lu2020phase} and \citet{mondelli2018fundamental} considered the problem with random matrix theory tools and discussed the weak recovery phase transition. More recently, \citet{maillard2022construction} connected these spectral methods to approximate message passing algorithms.

The traditional analyses of such spectral methods often rely on intricate random matrix theory arguments, requiring careful control of both the bulk distribution of eigenvalues and the behavior of isolated eigenvalues. We shall show that our framework provides a more direct route to analyzing these estimators. Specifically, consider the analysis of spectral estimators for matrices of the form
\begin{equation}
    A = \sum_{\dataidx=1}^n \mathcal{T}(y_\dataidx) \x_\dataidx \x_\dataidx^\top,
\end{equation}
where $\mathcal{T}: \RR \to \RR$ is a preprocessing function and the observations follow a GLM as $y_\dataidx = \outputfun(\frac{1}{d} \langle \wteacher, \x_\dataidx \rangle)$. The classical spectral method estimates the signal by computing the leading eigenvector of $A$. By the Courant-Fischer-Weyl min-max principle, this is equivalent to solving
\begin{equation}
    \lambda_{\mathrm{true}} = \min_{\w \in \mathcal{K}_{1,1}} \sum_{\dataidx=1}^n \mathcal{T}(y_\dataidx) \qty(\w^\top \x_\dataidx)^2 \,, \quad 
    \wminimizer = \argmin_{\w \in \mathcal{K}_{1,1}} \sum_{\dataidx=1}^n \mathcal{T}(y_\dataidx) \qty(\w^\top \x_\dataidx)^2 \,,
\end{equation}
where $\mathcal{K}_{1,1}$ denotes the unit sphere.

Consider the estimate $\wminimizer$ obtained from~\Cref{algo:gamp} and its associated eigenvalue estimate $\lambda_{\mathrm{amp}} = \langle \wminimizer, A \wminimizer \rangle/\norm{\wminimizer}_2^2$. In the high-dimensional limit where both the dimension $d$ and sample size $n$ grow to infinity at a fixed ratio $\alpha = n/d$, we prove that $\lambda_{\mathrm{amp}}$ converges almost surely to the true spectral estimate $\lambda_{\mathrm{true}}$. This convergence result relies on showing that the AMP algorithm converges to a fixed point that necessarily satisfies the first-order optimality conditions of the original spectral optimization problem.

Specifically, let $m = \langle \wteacher, \what \rangle / d$ denote the overlap of the solution with the teacher vector. In the high-dimensional limit, for $\rho = \lim_{d\to \infty}\norm{\wteacher}_2^2 / d = 1$, our main theorem proves that $m$ behaves as the solution of the following system of equations
\begin{align}
    m &= \frac{a(m,\tau)}{ \sqrt{\alpha a^2(m,\tau) + b(m,\tau)} } \,, & \, a(m, \tau) &= \EE{\qty(s - \frac{m h}{\sqrt{1 - m^2}} ) P(m,\tau)}  \,, \\ 
    \tau &= \frac{1}{ \alpha \sqrt{\alpha a^2(m,\tau) + b(m,\tau)} } \,, & \, b(m, \tau) &= \frac{1}{\tau^2} \EE{\qty(s - \frac{m h}{\sqrt{1 - m^2}} - P(m,\tau) )^2} \,,
\end{align}
\sloppy where the expectation is taken over $s, h \sim \mathcal{N}(0,1)$, $y \sim \pout{\cdot \mid s}$, and where $P(m,\tau) \defeq \mathcal{P}_{\tau \lossfun(y,\cdot) } \qty(s - \sfrac{m h}{\sqrt{1 - m^2}})$ with $\lossfun(y,z) = \mathcal{T}(y)z^2$.

These equations fully characterize the performance of both the spectral method and the AMP algorithm in the high-dimensional limit. 
This result provides a precise analytical description of the spectral estimator's performance without requiring random matrix theory tools.

Our analysis reveals that the algorithmic and statistical aspects of spectral estimation are intimately connected through the state evolution of AMP. 
In particular, the fixed point equations above allow us to precisely characterize the phase transition boundary between successful and unsuccessful recovery, as well as the asymptotic correlation achieved in the recovery phase.

\section{Outline of Proof}\label{sec:proof-sketch}

The key steps of the proof are three: a non-constructive lower bound derived from Gaussian comparison inequalities, a constructive upper bound obtained through an algorithmic procedure and a matching argument showing these bounds coincide. The algorithmic way to attain the lower bound is the message passing algorithm as the one presented in~\Cref{algo:gamp}.

\begin{algorithm}[t]
\DontPrintSemicolon
\KwIn{%
Feature Matrix $\datamat \in \RR^{n\times d}$, the vector of outputs $\y \in \RR^n$, initialization $\w^0 \in \RR^d$, set of functions $\{\outden[t], \inden[t]\}_{t\geq 0}$
}
\KwOut{Value of $\w \in \RR^d$ and $\p \in \RR^n$ at termination}

\BlankLine
\While{$\lnot$ Termination Condition}{
    \tcp{Output Node Update}
    \(b^t := \frac{1}{d} \sum_{\dataidx = 1}^n \partial \outden[t] \qty(p_\dataidx^t, y_\dataidx)\)\;
    
    \(\p^t = \frac{1}{\sqrt{d}}
    \datamat \inden[t] \qty( \w^t ) - b^t
    \outden[t-1] \qty(\p^{t-1}, \y)\)  \;
    
    \tcp{Input Node Update}
    \(c^t := \frac{1}{d} \sum_{\dimidx = 1}^d \partial \inden[t] \qty(w_\dimidx^t)\)\;
    
    \(\w^{t+1} = \frac{1}{\sqrt{d}}
    \datamat^\top \outden[t] \qty(\p^t, \y) - c^t 
    \inden[t] \qty( \w^t )\)\;
}
\caption{
General Approximate Message Passing (GAMP). 
}
\label{algo:gamp}
\end{algorithm}

The first step is to derive the lower bound on the minimum achievable risk in the high-dimensional limit. This is the following result
\begin{proposition}\label{thm:lower_b}
Let $\minriskfun$ be the minimal value of the risk defined in \cref{eq:min-risk-limit-def} and $\minriskfun_d$ be the minimum value defined in \cref{eq:minimum-risk}. Then, 
in the high-dimensional limit where $n,d\to \infty$ and $\alpha = n/d$ we have that
for every $\varepsilon > 0$
\begin{equation}
    \Pr[
        \liminf_{d \rightarrow \infty} \minriskfun_d \geq \minriskfun - \varepsilon
    ] = 1 \,.
\end{equation}
\end{proposition}

This proposition establishes that the limiting risk $\minriskfun$ serves as an asymptotic lower bound for the finite-dimensional minimum $\minriskfun_d$. 
This lower bound admits an explicit characterization in terms of Moreau envelopes of the loss and regularization functions that is stated in full generality in~\Cref{sec:app:lower-bound}.

The next step in the proof is to use the scalar parmeters found from the minimisation of~\cref{eq:min-risk-limit-def} and use them to build an initializiation for~\Cref{algo:gamp}. We show that under our assumptions this implies the termination of the algorithm in the following 

\begin{lemma}[Convergence of GAMP algorithm]\label{lem:stab}
Let $(m^\star, q^\star, \tau^\star, \kappa^\star, \nu^\star, \eta^\star) \in \mathcal{S}^\star$ defined in~\cref{eq:def-s-star}, denote a set of parameters satisfying \Cref{ass:weak_convexity,ass:stab}.
Consider \Cref{algo:gamp} initialized as
\begin{equation}\label{eq:init-gamp}
    \w^0 = \frac{m^\star}{\rho} \wteacher + \sqrt{q^\star - \frac{(m^\star)^2}{\rho}} \xib \,,
\end{equation}
where $\xib \sim \mathcal{N}(\boldsymbol{0}, \operatorname{Id}_d)$ independently from all other quantities. 
We will indicate as $\w^t$ the iterates of \Cref{algo:gamp} with non linearities $\{g^t=\outden[\star],f^t = \inden[\star]\}_{t\geq 0}$, defined in~\cref{eq:output-input-denoisers-amp}. 
Then there exist constants $C_1, C_2 > 0$ and $0 < \rho_1,\rho_2  < 1$ such that:
\begin{equation}
    \mathcal{E}_{w,\rho_1,d} \coloneqq \qty{
        \frac{1}{\sqrt{d}} \norm{\w^{t+1} - \w^{t}}_2 \leq C_1 {\rho_1}^t
    } \,, \qquad 
    \mathcal{E}_{p,\rho_2,d} \coloneqq \qty{ 
        \frac{1}{\sqrt{n}} \norm{\p^{t+1} - \p^{t}}_2 \leq  C_2 {\rho_2}^t
    }\,, 
\end{equation}
hold almost surely as $n,d \rightarrow \infty$ i.e:
\begin{equation}
\Pr[\cap_{d\geq d_0}\mathcal{E}_{w,\rho_1,d}] \rightarrow 1, \, \quad \Pr[\cap_{d\geq d_0}\mathcal{E}_{p,\rho_2,d}]\rightarrow 1,
\end{equation}
as $d_0 \rightarrow \infty$.
\end{lemma}

The previous lemma is proven by considering the reduction of the message passing algorithm to its state evolution dynamics and then showing the contractivity of the evolution. 
Because of the separable form of the regularization term we analyze an equivalent version of the algorithm where the dependence on the norm is exchanged by a rescaling by a constant. 

Finally we show that the fixed points of~\Cref{algo:gamp} obtained with the previous initialization reach the lower bound on the energy thus proving that it is attainable.

\begin{proposition}\label{prop:physicality-fixed-point}
    Let $(m^\star, q^\star, \tau^\star, \kappa^\star, \nu^\star, \eta^\star) \in \mathcal{S}^\star$ denote a set of parameters satisfying Assumptions \ref{ass:weak_convexity} and \ref{ass:stab}.
    Then for the sequence of iterates obtained by applying~\Cref{algo:gamp} starting from initialization of~\cref{eq:init-gamp} we have
    \begin{equation}
        \lim_{t \to \infty} \lim_{d \to \infty}\mathcal{A}_d(\hat{\w}^t) = \minriskfun \,,
    \end{equation}
    almost surely.
\end{proposition}

We have thus proved that the estimate returned by the message passing algorithm reaches the lower-bound given from the Gaussian comparison theorem, ending the main part of the proof.

Our proof technique's originality lies in synthesizing two established methodologies in high-dimensional statistics. 
While each approach has been studied separately, their combination yields a framework that mirrors the intuition from statistical mechanics. Specifically, our approach first derives equations for the order parameters through comparison inequalities, analogous to the replica method, and then verifies their physical realizability through the stability analysis of the corresponding dynamics. 

The methodology developed in this paper opens several promising research directions. 
A natural extension would be to problems exhibiting replica symmetry breaking, where the current Gaussian comparison bounds would need to be generalized to account for multiple pure states. 
Such an extension could provide insights into problems like sparse PCA or community detection in the hard phase. 
Furthermore, our proof technique suggests a general recipe for analyzing non-convex optimization landscapes: first establishing information-theoretic limits through comparison inequalities, then constructing explicit algorithms that achieve these bounds. 

\section{Preliminaries}\label{sec:app:preliminaries}

This section will be devoted to the statement of definitions, preliminary results and theorems that will be used in the following.
We will use the following definition extensively.
\begin{definition}[Pseudo-Lipschitz function]\label{def:pseudo-lipshitz-fucntion}
A function $f \colon \RR \to \RR$ is said to be \emph{pseudo-Lipschitz of order $k$} if there exists a constant $L$ such that for any $x,y \in \RR$, 
\begin{equation}
    \abs{f(x) - f(y)} \leq 
    L \qty(
        1 
        + \abs{x}^{k-1} 
        + \abs{y}^{k-1}
    )
    \abs{x - y} \,.
\end{equation}
\end{definition}

We also define the following two proprierties of a function
\begin{definition}[Prox-Coercive function]
A function $f$ is said to be prox-coercive if for any $r, \delta \in \RR \mapsto f(r) + \frac{(\delta- r)^2}{2}$ is coercive in $r$; \textit{i.e.}
\begin{equation}
    \lim _{|r| \rightarrow \infty} f(r)+\frac{(\delta-r)^2}{2}=\infty \,.
\end{equation}
\end{definition}

\begin{definition}[Locally lower bounded]\label{def:locally-lower-bounded}
A function $f$ is said to be locally lower bounded if, for any point $r\in\RR$, there exists a neighborhood $U\subset\RR$ of $r$ such that
\begin{equation}
    \inf_U f > - \infty\,.
\end{equation}    
\end{definition}

\begin{definition}[Weakly convex function]\label{def:weakly-convex-definition}
Fix $V_0\in [0,\infty)$. A function $f$ is said to be $V_0$-weakly convex (resp. strictly $V_0$-weakly convex) if the function $r \mapsto V_0f(r)+\frac{r^2}{2}$ is convex (resp. strictly convex).
A function $f$ is said to be weakly convex if it is $V_0$-weakly convex for some $V_0\in (0,\infty)$.
\end{definition}

\begin{remark}
From this definition three remarks are in order
\begin{itemize}
    \item $f$ is convex if and only if it is $V_0$-weakly convex for any $V_0\in [0,\infty)$.
    \item For a given $V_0\in [0,\infty)$ $f$ is $V_0$-weakly convex if and only if $V_0f$ is $1$-weakly convex.
    \item If $f$ is $V_0$ weakly convex for some $V_0>0$, then $f$ is strictly $V$-weakly convex for $V\in[0,V_0)$.
\end{itemize}
\end{remark}

To require that $f$ is weakly convex is much less restrictive than the name might suggest. Informally, it can be understood as having a lower bounded second derivative almost everywhere.

\begin{remark}
We additionally note that the notable example of the negative perceptron, where the considered function is the negative step function is not considered in our framework.
\end{remark}

\begin{definition}[Effective Domain of Proximal Operator]\label{def:effective-domain-prox}
For a function $f : \mathbb{R} \to \mathbb{R}$, we define $E_f := \{ \delta \in \mathbb{R} : \mathcal{P}_f(\delta) \neq \emptyset \}$ as the set of points for which the proximal is non-empty.
\end{definition}

For analyzing the regularity properties of proximal selections, we introduce two auxiliary functions that capture the behavior of the modified function $r \mapsto f(r) + \tfrac{r^2}{2}$:

\begin{definition}[Chord Slope Functions]\label{def:app:si}
For a function $f : \mathbb{R} \to \mathbb{R}$, we define $s : \mathbb{R} \to \mathbb{R} \cup \{\infty\}$ and $i : \mathbb{R} \to \mathbb{R} \cup \{-\infty\}$ by
\begin{equation}
s_f(r_0) := \sup_{r < r_0} \left[ \frac{f(r) - f(r_0)}{r - r_0} + \frac{r + r_0}{2} \right] \,, \quad 
i_f(r_0) := \inf_{r > r_0} \left[ \frac{f(r) - f(r_0)}{r - r_0} + \frac{r + r_0}{2} \right] \,.
\end{equation}
\end{definition}

\begin{remark}
The functions $s_f(r_0)$ and $i_f(r_0)$ represent the supremum and infimum, respectively, of the slopes of the chords at $r_0$ of the function $r \mapsto f(r) + \tfrac{r^2}{2}$.  
When $f$ is $1$-weakly convex (i.e., $r \mapsto f(r) + \tfrac{r^2}{2}$ is convex), we have
$s = f'_{-} + \mathrm{id}$ and
$i = f'_{+} + \mathrm{id}$
where $f'_{-}$ and $f'_{+}$ denote the left and right derivatives of $f$, respectively.
\end{remark}

\subsection{Useful Lemmas about Moreau Envelopes and Proximal Selections}

The first lemma that we need to use in the following has already appeared as~\citep[Lemma~5]{loureiro2021learning} but the proof relied on convexity. We thus extend it to the setting considered in this manuscript.

\begin{lemma}[Pseudo-Lipschitz property of Moreau envelopes]\label{lem:pseudo-lip-moreau}
    Suppose that $\lossfun : \RR \times \RR  \rightarrow \RR, \regfun: \RR \rightarrow \RR$ are pseudo-Lipschitz of order $k$, and such that $E_{\lossfun(y,\cdot)}=\RR$, $E_{\regfun}=\RR$ for all $y\in \RR$, then $(y,\omega) \mapsto \mathcal{M}_{\lossfun(y, \cdot)}(\omega)$, $\omega \mapsto \mathcal{M}_{\regfun(\cdot)}(\omega)$ are pseudo-Lipschitz.
\end{lemma}

\begin{proof}[\Cref{lem:pseudo-lip-moreau}]
We prove the result for $\mathcal{M}_{\lossfun(y,\cdot)}$; the argument for $\mathcal{M}_{\regfun(\cdot)}$ is analogous. 

Let $(y_1,\omega_1), (y_2,\omega_2) \in \RR^2$ be arbitrary and let $z_1, z_2$ denote any of the corresponding extremizers of $\mathcal{M}_{\lossfun(y_1, \cdot)}(\omega_1)$ and $\mathcal{M}_{\lossfun(y_2, \cdot)}(\omega_2)$ respectively. 
Define:
\begin{align}
   \Phi_{y} (z,\omega) \coloneqq 
   \lossfun (y,z) + \frac{1}{2} \abs{z - \omega}^2 \,.
\end{align}
Recall that:
\begin{align}
    \mathcal{M}_{\lossfun (y_i,\cdot)}(z_i) = \inf_{z \in \mathbb{R}} \Phi_{y_i} (z,\omega_i),
\end{align}
for $i=1,2$.

In particular,
\begin{align}
    \lossfun(y_1,z_1) + 
    \frac{1}{2} \abs{z_1 - \omega_1}^2 = \mathcal{M}_{\lossfun (y_1,\cdot)}(\omega_1) \leq \lossfun(y_1,z_2) + 
    \frac{1}{2} \abs{z_2 - \omega_1}^2 \,, \label{eq:app:optboundz1} \\
    \lossfun(y_2,z_2) + 
    \frac{1}{2} \abs{z_2 - \omega_2}^2 = \mathcal{M}_{ \lossfun (y_2,\cdot)}(\omega_2) \leq \lossfun(y_2,z_1) + 
    \frac{1}{2} \abs{z_1 - \omega_2}^2 \,. \label{eq:app:optboundz2}
\end{align}

Hence we deduce :

\begin{align}
    \mathcal{M}_{\lossfun (y_1,\cdot)}(\omega_1) - \mathcal{M}_{ \lossfun (y_2,\cdot)}(\omega_2) \leq \lossfun(y_1,z_2) + 
    \frac{1}{2} \abs{z_2 - \omega_1}^2 - \lossfun(y_2,z_2) -
    \frac{1}{2} \abs{z_2 - \omega_2}^2 \,,
    \label{eq:app:optupbound} \\
    \mathcal{M}_{\lossfun (y_1,\cdot)}(\omega_1) - \mathcal{M}_{ \lossfun (y_2,\cdot)}(\omega_2) \geq \lossfun(y_1,z_1) + 
    \frac{1}{2} \abs{z_1 - \omega_1}^2 -\lossfun(y_2,z_1) - 
    \frac{1}{2} \abs{z_1 - \omega_2}^2 \,.
    \label{eq:app:optlowbound}
\end{align}

Furthermore, again by the optimality of $z_i$, we have $\Phi_{y_i} (z_i,\omega_i) \leq \Phi_{y_i} (0,\omega_i)$ \emph{i.e.}:
\begin{equation}
   \lossfun(y_i, z_i) + \frac{1}{2}\abs{\omega_i - z_i}^2 \leq  
   \lossfun(y_i,0) + \frac{1}{2}\abs{\omega_i}^2 \,.
\end{equation}
for $i = 1, 2$.

Since $\lossfun$ is bounded below by~\Cref{ass:lossregfun} by some constant $\mu_L$, we obtain
\begin{equation}\label{eq:app:norm_bound}
    \abs{z_i-\omega_i} \leq \sqrt{2 (\lossfun(y_i,0)-\mu_L)+ \abs{\omega_i}^2} \,,
\end{equation}
for $i=1,2$.

Using the pseudo-Lipschitzness of $\lossfun$ yields:
\begin{equation}\label{eq:app:norm_bound_2}
    \abs{z_i-\omega_i} \leq \sqrt{c_1+c_2\abs{y_i}^k+ \abs{\omega_i}^2} \,,
\end{equation}
for some constants $c_1, c_2 > 0$.
Using the triangular inequality, $z_i$ is also bounded by a polynomial in $|y_i|,|\omega_i|$.

Now let $i,j=1,2$ be distinct. Using the polynomial bounds above, we deduce that 
\begin{align}
    \left|\frac{1}{2}|z_i-\omega_j|^2 - \frac{1}{2}|z_i-\omega_i|^2\right| &= \frac{1}{2}|\omega_j-\omega_i||2z_i-\omega_j-\omega_i| \leq \frac{1}{2}|\omega_j-\omega_i|(2|z_i|+|\omega_j|+|\omega_i|) \,,
\end{align}
is bounded by a polynomial in $|y_1|, |y_2|, |\omega_1|, |\omega_2|$ times $|\omega_1-\omega_2|$.

All that remains now to conclude using \cref{eq:app:optupbound} and \cref{eq:app:optlowbound} is to deal with the terms $\lossfun(y_j,z_i)-\lossfun(y_i,z_i)$.

Next, by the pseudo-Lipschitzness of order $k_L$ of $\lossfun$ we have:
\begin{align} 
   \lossfun(y_1,z_2) - \lossfun(y_2,z_2) & \leq L(1+2\abs{z_2}^{k_L-1}+\abs{y_1}^{k_L-1}+ \abs{y_2}^{k_L-1}) \abs{y_1-y_2}\\
    \lossfun(y_2,z_1) - \lossfun(y_1,z_1) & \leq
      L(1+2\abs{z_1}^{k_L-1}+\abs{y_1}^{k_L-1}+ \abs{y_2}^{k_L-1}) \abs{y_1-y_2}
\end{align}

Similarly, since we can bound the terms $|z_1|^{k_L-1}, |z_2|^{k_L-1}$ polynomialy in $|y_1|, |y_2|, |\omega_1|, |\omega_2|$, we obtain the desired result.
Similarly, we can obtain the corresponding result for $\regfun$.
\end{proof}

\begin{lemma}[Coercivity of Moreau Envelopes]\label{lem:app:coercivity-moreau}
For any coercive function $f : \RR \to \RR$ lower bounded by $C$, the Moreau envelope $\mathcal{M}_{f(\cdot)}$ is coercive.
\end{lemma}

\begin{proof}[\Cref{lem:app:coercivity-moreau}]
By the definition of Moreau Envelope form~\cref{eq:main:moreau-definition}, we have:
\begin{equation}
    \mathcal{M}_{f(\cdot)}(x) = \inf_{y \in \RR} \left\{ f(y) + \frac{1}{2} \abs{x - y}^2 \right\} \,.
\end{equation}

The coercivity of $f$ implies that for any $M>0$, $\exists R(M) > 0$ such that:
\begin{equation}
    \forall \abs{x} > R(M) \implies f(x) > M.
\end{equation}

Now, set $\tilde{R} = R(M)+\sqrt{2(M+|C|)}$. Suppose that $\abs{x} > \tilde{R}$. Then, for any $y \in \RR$ with $\abs{y} > R(M)$, we have:
\begin{equation}
   f(y) + \frac{1}{2} \abs{x - y}^2 \geq f(y) > M \,,
\end{equation}
while for any $y \in \RR$ with $\abs{y} \leq R(M)$:
\begin{align}
    f(y) + \frac{1}{2} \abs{x - y}^2 &\geq \frac{1}{2} \abs{x - y}^2 - C \geq \frac{1}{2} (\abs{x} - \abs{y})^2-C \geq \frac{1}{2} (\abs{x} - R(M))^2-C > \frac{1}{2} (\sqrt{2(M+|C|})^2-C \geq M \,,
\end{align}
where the last inequality follows since $\abs{x} > \tilde{R} = R(M)+\sqrt{2M}$. Therefore, for any $y \in \RR$, we have that $f(y) + \frac{1}{2} \abs{x - y}^2 > M$. Taking the infimum over $y$, we conclude that $\abs{x} > \tilde{R} \implies \mathcal{M}_{ f(\cdot)}(x) \geq M$, establishing the coercivity of $\mathcal{M}_{ f(\cdot)}$.
\end{proof}

The other two tools that are needed in the subsequent analysis concern the  characterization of proximal selections under weak convexity conditions. 
The following theorem is the main technical tool that allows us to translate weak convexity assumptions on the loss and regularization functions into Lipschitz continuity properties of their proximal operators. 

\begin{theorem}[Main Characterization of Proximal Selections]\label{thm:main_proximal_characterization}
Suppose $f:\RR \to \RR$ is lower semi-continuous and prox-coercive. 
Then the proximal operator $\mathcal{P}_f$ of $f$ is never empty-valued, that is, $E_f = \mathbb{R}$. When this condition holds, 
any proximal selection $r_{\min}$ of $f$ is non-decreasing, 
almost everywhere differentiable, and the following equivalences are satisfied:
\begin{equation}
    (i) \qquad r_{\min} \in C^0(\mathbb{R}, \mathbb{R}) \iff f \text{ is strictly } 1\text{-weakly convex},
\end{equation}
\begin{equation}
    (ii) \qquad r_{\min} \text{ is } 1\text{-Lipschitz continuous over } \mathbb{R} \iff f \text{ is convex},
\end{equation}
and for any $L \in (1, \infty)$,
\begin{equation}
    (iii) \qquad r_{\min} \text{ is } L\text{-Lipschitz continuous over } \mathbb{R} \iff f \text{ is } \frac{L}{L-1}\text{-weakly convex}.
\end{equation}

Furthermore, when any of these conditions hold, $r_{\min}$ is uniquely determined by the set equality
\begin{equation}
\forall \delta \in \mathbb{R}, \quad \mathcal{P}_f(\delta) = \{r_{\min}(\delta)\}\,.
\end{equation}

In this case, $r_{\min}$ is onto, and for $r_0 \in \mathbb{R}$, the interval over which $r_{\min}$ takes the constant value $r_0$ is given by
\begin{equation}
[r_0 + f'_-(r_0), r_0 + f'_+(r_0)] \,.
\end{equation}

In particular, $r_{\min}$ is increasing if and only if $f$ is differentiable.
\end{theorem}

The following corollary establishes that~\Cref{ass:weak_convexity} guarantees the existence and uniqueness of Lipschitz continuous proximal selections for both the loss and regularization functions appearing in~\cref{eq:output-input-denoisers-amp}.

\begin{corollary}[Scaled Proximal Operators]\label{cor:scaled_proximal}
Let $f : \mathbb{R} \to \mathbb{R}$ be $V_0$-weakly convex for some $V_0 > 0$. For any $V\in (0,V_0)$, $Vf$ is $V_0/V$-weakly convex and $E_{Vf}=\RR$. Furthermore, with $L=\frac{V_0}{V_0-V}$, there exists a unique $L$-Lipschitz continuous selection $\tilde{r} : \mathbb{R} \to \mathbb{R}$ such that 
$$\forall \delta\in \RR, \: \tilde{r}(\delta) \in \argmin_{r \in \mathbb{R}} \left\{\frac{(\delta-r)^2}{2} + Vf(r)\right\}.$$

Moreover under~\Cref{ass:weak_convexity} the function $\lossfun(y,\cdot)$ (respectively $\regfun$) define a unique $L_L$-Lipschitz continuous (respectively $L_R$-Lipschitz continuous) selection of their respective proximal operators.
\end{corollary}

The proofs of~\Cref{thm:main_proximal_characterization,cor:scaled_proximal} are deferred to~\Cref{sec:app:from-weak-convex-to-lipshitz}.

\subsection{Technical Framework}

Our analysis relies on two theoretical frameworks: Generalized Approximate Message Passing (GAMP) and the Gaussian Min-max Theorem (GMT). 

\vspace{1em}
GAMP is a class of iterative algorithms characterized by coupled iterations, as in~\Cref{algo:gamp}, which we defined from a set of separable functions $\{\outden[t], \inden[t]\}_{t\geq 0}$ and their associated Onsager correction terms $b^t, c^t$ defined as
\begin{equation}\label{eq:onsager-terms}
    b^t := \frac{1}{d} \sum_{\dataidx = 1}^n \partial \outden[t] \qty(p_i^t, y_i) \,, \quad c^t := \frac{1}{d} \sum_{\dimidx = 1}^d \partial \inden[t] \qty(w_j^t) \,,
\end{equation}
where derivatives are taken with respect to the first argument. 

In the high-dimensional limit of \Cref{ass:high-dimensional-limit}, the algorithm admits a dimensional reduction through its State Evolution equations \citep{javanmard_montanari_gamp_2013,berthier2020state,feng2022unifying,gerbelot2023graph}. 
These equations track one-dimensional statistics of the iterates through
\begin{align}
    \beta_t &= \frac{\mathbb{E}\qty[ \tilde{w} \inden[t]( \mu_t \tilde{w} + u^t ) ]}{\mathbb{E}\qty[(\tilde{w})^2]} \,, \label{eq:beta-def} \\
    \Omega_{s,t} &= \mathbb{E}\qty[ \inden[s](\mu_s \tilde{w} + u^{s}) \inden[t](\mu_t \tilde{w} + u^{t}) ] - \mathbb{E}\qty[(\tilde{w})^2] \beta_t \beta_s \,, \label{eq:omega-def} \\
    \mu_{t+1} &= \alpha \mathbb{E}\qty[ v^\star \outden[t](\beta_t v^\star + v^t) ] \,, \label{eq:mu-def} \\
    \Sigma_{s+1,t+1} &= \alpha \mathbb{E}\qty[ \outden[s](\beta_s v^\star + v^s) \outden[t](\beta_t v^\star + v^t) ] \,, \label{eq:sigma-def}
\end{align}
where $(u^1, \dots u^t) \sim \mathcal{N}(\boldsymbol{0}, \boldsymbol{\Sigma})$ and $(v^0, \dots v^t) \sim \mathcal{N}(\boldsymbol{0}, \boldsymbol{\Omega})$ are independent Gaussian vectors, $v^\star \sim \mathcal{N}(0,1)$ and $\tilde{w}$ comes from~\Cref{ass:teacher-channel}.
The previous quantities are defined recursively from the initial condition $\mu_0 = \mathbb{E}\qty[\tilde{w} \outden[0](u^0)]$ and $\Sigma_{0,0} = \mathbb{E}\qty[ \outden[0](u^0)^2 ]$ where $u_0$ is distributed as the initialisation of the problem.

The validity of the state evolution equations is then provided by the following result:

\begin{theorem}[\citet{javanmard_montanari_gamp_2013,berthier2020state,gerbelot2023graph}]\label{lem:state-evolution-general-covariates}
Consider~\Cref{algo:gamp} with pseudo-Lipschitz functions $\{\inden[t],\outden[t]\}_{t \geq 0}$ and initialization $\w^0$ such that the joint empirical distribution of $(\w^0, \wteacher)$ converges almost surely to a probability measure $P_0$ on $\mathbb{R}^2$ with bounded $k$-th moments for some $k \geq 2$. Let $u^0$ and $\tilde{w}$ denote random variables with law $P_0$, and define the initial values for the recursion as
\begin{equation}\label{eq:state-evolution-init}
    \mu_0 = \mathbb{E}\qty[\tilde{w} \outden[0](u^0)] \,, \quad \Sigma_{0,0} = \mathbb{E}\qty[ \outden[0](u^0)^2 ] \,.
\end{equation}

Then, under~\Cref{ass:high-dimensional-limit}, for any finite time $t \in \N$ and any pseudo-Lipschitz function $\psi : \RR^{t+1} \rightarrow \RR$, we have
\begin{align}
    \frac{1}{d} \sum_{\dimidx = 1}^d \psi(\tilde{w}_\dimidx, w^{1}_\dimidx, \dots, w^{t}_\dimidx) - 
    \mathbb{E}\qty[\psi(\tilde{w}, \mu_1 \tilde{w} + u^1, \dots, \mu_t \tilde{w} + u^t) ] & \xrightarrow{a.s.} 0 \,, \label{eq:state-evolution-w} \\ 
    \frac{1}{n} \sum_{\dataidx = 1}^n \psi(p^\star_\dataidx, p^{0}_\dataidx, \dots, p^{t}_\dataidx) - 
    \mathbb{E}\qty[\psi(v^\star, \beta_0 v^\star + v^0, \dots, \beta_t v^\star + v^t ) ] & \xrightarrow{a.s.} 0 \,, \label{eq:state-evolution-p}
\end{align}
where $(u^1, \dots u^t)$ and $(v^0, \dots v^t)$ denote zero-mean Gaussian vectors with 
$(u^1, \dots u^t) \sim \mathcal{N}(\boldsymbol{0}, \boldsymbol{\Sigma})$ and $(v^0, \dots v^t) \sim \mathcal{N}(\boldsymbol{0}, \boldsymbol{\Omega})$
where $\boldsymbol{\Sigma}, \boldsymbol{\Omega} \in \RR^{{(t + 1)} \times {(t + 1)}}$ are defined recursively by~\cref{eq:beta-def,eq:omega-def,eq:mu-def,eq:sigma-def} 
with initial conditions~\cref{eq:state-evolution-init} and with $p_\mu^\star = \x_\mu^\top \wteacher / \sqrt{d}$.
\end{theorem}

For any value $t \in \N$ we also note that one can estimate the value of the minimizer with $\what^t = \inden[t](\w^t)$ \citep{feng2022unifying}, where $\w^t$ is the iterate from~\Cref{algo:gamp}.

\vspace{1em}
Complementing the previous is Gordon's Gaussian comparison inequality~\citep{gordon_1988}, that allows us to study challenging non-convex optimization problems by comparing them to simpler auxiliary problems involving only vectorial Gaussian processes. 

\begin{theorem}[GMT \cite{gordon_1988}]\label{thm:gmt}
Let \(\boldsymbol{G} \in \RR^{m \times n}\) be an i.i.d. standard normal matrix and \(\mathbf{g} \in \RR^m\), \(\mathbf{h} \in \RR^n\) two i.i.d. standard normal vectors independent of one another. Let \(\mathcal{S}_{\w}\), \(\mathcal{S}_{\boldsymbol{u}}\) be two compact sets such that \(\mathcal{S}_{\w} \subset \RR^n\) and \(\mathcal{S}_{\boldsymbol{u}} \subset \RR^m\). Consider the two following optimization problems for any continuous \(\psi\) on \(\mathcal{S}_{\w} \times \mathcal{S}_{\boldsymbol{u}}\)
\begin{align}
    \mathbf{C}(\boldsymbol{G}) &:= \min_{\w \in \mathcal{S}_{\w}} \max _{\boldsymbol{u} \in \mathcal{S}_{\boldsymbol{u}}} 
    \boldsymbol{u}^{\top} \boldsymbol{G} \w + 
    \psi(\w, \boldsymbol{u}) \label{eq:gmt:primary-problem} \\
    \mathcal{C}(\mathbf{g}, \mathbf{h}) &:= \min_{\w \in \mathcal{S}_{\w}} \max _{\boldsymbol{u} \in \mathcal{S}_{\boldsymbol{u}}} 
    \|\w\|_2 \mathbf{g}^{\top} \boldsymbol{u} + 
    \|\boldsymbol{u}\|_2 \mathbf{h}^{\top} \w + 
    \psi(\w, \boldsymbol{u}) \label{eq:gmt:auxiliary-problem}
\end{align}
Then for all \(c\in \RR\) we have
\begin{equation}
    \Pr[\mathbf{C}(\boldsymbol{G})<c] \le 2 \Pr[\mathcal{C}(\mathbf{g}, \mathbf{h}) \le c]
\end{equation}
\end{theorem}

Under concentration of $\mathcal{C}(\mathbf{g}, \mathbf{h})$, the above theorem yields a high-probability lower bound on $\mathbf{C}(\boldsymbol{G})$. This is desirable since unlike $\mathbf{C}(\boldsymbol{G})$ which 
involves a Gaussian matrix $\boldsymbol{G} \in \RR^{m \times n}$, its auxiliary counterpart depends only on independent Gaussian vectors $\mathbf{g} \in \RR^m, \mathbf{h} \in \RR^n$ and is often more tractable to analyze. 

\section{Proofs for the Gordon Lower Bound}\label{sec:app:lower-bound}

The main objective of this section is to rewrite the minimization problem \cref{eq:minimum-risk} in a way amenable to the application of~\Cref{thm:gmt} with the end goal of proving~\Cref{thm:lower_b}.
Specifically we start from the following problem
\begin{equation}\label{eq:app:risk-def}
    \inf_{\w \in \mathcal{K}_{a,b}} \frac{1}{d} \qty[ 
        \sum_{\dataidx=1}^n 
        \lossfun\qty(y_\dataidx, \frac{\w^\top \x_\dataidx}{\sqrt{d}} ) + 
        \sum_{\dimidx=1}^{d} \regfun(\w_\dimidx) 
    ] \,,
\end{equation}
which is equivalent to
\begin{equation}
    \inf_{\w \in \mathcal{K}_{a,b}, \z \in \RR^n} \frac{1}{d}\left[
        \sum_{\dataidx=1}^n \lossfun(y_{\dataidx}, z_{\dataidx}) + 
        \sum_{\dimidx=1}^d \regfun(w_\dimidx)
    \right] \, 
    \text{ subject to }
    z_{\dataidx} = \frac{\w^\top \x_{\dataidx}}{\sqrt{d}} \,, \quad \forall \dataidx \in [n] \,.
\end{equation}
We can introduce the Lagrange multiplier $\f \in \RR^n$ to obtain the following form of the problem
\begin{equation}\label{eq:app:risk-def-duality}
    \minriskfun_d = 
    \inf_{\w \in \mathcal{K}_{a,b}, \z\in\RR^n} 
    \sup_{\f \in \RR^n}
    \frac{1}{d}
    \qty[ 
        \frac{1}{\sqrt{d}} \f^\top \datamat \w
        - \f^\top \z 
        + \sum_{\dataidx=1}^{n} \lossfun\qty(y_\dataidx, z_\dataidx)
        + \sum_{\dimidx=1}^{d} \regfun\qty(w_\dimidx) 
    ] \,.
\end{equation}

\begin{proposition}\label{prop:app:compactness-high-dim}
Defining $\mathcal{S}_{C,d} = \qty{\vv \in \RR^d \mid \norm{\vv}_2 \leq C \sqrt{d}}$. The optimal values $\what$, $\hat{\f}$, $\hat{\z}$ of \cref{eq:app:risk-def-duality} satisfy:
\begin{equation}
    \Pr[\cap_{d \geq d_0} \left[\what \in \mathcal{S}_{C,d} \cap \mathcal{S}_{a,b}\right]] \xrightarrow[d_0\to \infty]{} 1 \,, \quad 
    \Pr[\cap_{d \geq d_0} \left[\hat{\z} \in \mathcal{S}_{C,n} \right]] \xrightarrow[d_0\to \infty]{} 1 \,, \quad 
    \Pr[\cap_{d \geq d_0} \left[\hat{\f} \in \mathcal{S}_{C,n}\right]] \xrightarrow[d_0\to \infty]{} 1 \,, 
\end{equation}
for a positive constant $C$ independent of $d,n$.
\end{proposition}

\begin{proof}[\Cref{prop:app:compactness-high-dim}]
To start we construct a feasible comparison point $\bar{\w} \in \mathcal{K}_{a,b}$.
We define $\bar{\w}_i = \sqrt{a}$ for all $i \in [d]$, so that $\|\bar{\w}\|_2^2/d = a$.
By the definition of $\what$ from~\cref{eq:minimiser-risk} we have that $\riskfun_d(\what) \leq \riskfun_d(\bar{\w})$.

We define the following random variable
\begin{equation}
    Z_\dataidx \defeq \frac{\bar{w}^\top x_\dataidx}{\sqrt{d}} = \frac{\sqrt{a}}{\sqrt{d}}\sum_{i=1}^d x_{\dataidx,i} \sim \mathcal{N}(0, a) \,.
\end{equation}

By pseudo-Lipschitzness of $\lossfun$ and boundedness of $y$ (from~\Cref{ass:teacher-channel}), the random variables $\lossfun(y_\dataidx, Z_\dataidx)$ are independent across $\dataidx$ (since $x_\dataidx$ are independent) and sub-exponential. 
Their expectation
\begin{equation}
    C_L(a) \defeq \mathbb{E}_{x}[\lossfun(y, Z)] < \infty
\end{equation}
is finite, where $\x \sim \mathcal{N}(0, I_d)$, $s = \langle \wteacher, x\rangle/\sqrt{d}$, $y \sim P_{\mathrm{out}}(\cdot | s)$, and $Z = \sqrt{a}\sum_{i=1}^d x_i/\sqrt{d}$ are the coupled random variables induced by $\x$.

By Bernstein's inequality \citep[Theorem~2.9.1]{vershynin2018high}, there exists a constant $C^\prime > 0$ such that:
\begin{equation}
    \frac{1}{n}\sum_{\dataidx=1}^n \lossfun(y_\dataidx, Z_\dataidx) \leq C_L(a) + C^\prime \,,
\end{equation}
with probability at least $1 - 2\exp(-cn)$ for some absolute constant $c > 0$.

Since $\lossfun$ is bounded below by some $\mu_L$, we have
\begin{equation}\label{eq:app:chain-ineq}
    \alpha \mu_L + \frac{1}{d} \sum_{\dimidx=1}^d \regfun\left(\hat{w}_\dimidx\right) 
    \leq \alpha(C_L(a) + C') + \regfun(\sqrt{a}) \,,
\end{equation}
with probability at least $1 - 2\exp(-cn)$.

Under~\Cref{ass:lossregfun}, using the quadratic lower bound $\regfun(w) \geq C_R w^2$ for $|w| \geq W_0$ we have that~\cref{eq:app:chain-ineq} implies:
\begin{equation}
    \frac{1}{d} \sum_{\dimidx=1}^d \hat{w}_\dimidx^2 \leq C\left(\alpha, a, \mu_L, \regfun(\sqrt{a}), C_R, W_0\right),
\end{equation}
with probability at least $1 - 2\exp(-cn)$, where the constant is independent of $d,n$. Therefore $\frac{1}{d}\|\what\|_2^2 \leq C_w$ with probability at least $1 - 2\exp(-cn)$ for some $C_w > 0$.

To bound $\hat{\z}$ we recall that $\hat{\z} = \frac{1}{\sqrt{d}} \datamat \what$. We start by applying \citep[Theorem~4.4.5]{vershynin2018high} with $t = \sqrt{d}$, which gives
\begin{equation}\label{eq:bound-operator-norm}
    \frac{1}{\sqrt{d}}\norm{\datamat}_{\mathrm{op}} \leq K (2 + \sqrt{\alpha}) =: C_\mathrm{op}\,,
\end{equation}
with probability at least $1 - 2\exp(-d)$ for a positive constant $K$. We can thus bound the following
\begin{equation}
    \frac{1}{n}\norm{\hat{\z}}_2^2 \leq 
    \frac{1}{d}\norm{\datamat}_{\mathrm{op}}^2 \frac{1}{\alpha} \frac{1}{d}\norm{\what}_2^2 \leq 
    \frac{ C_\mathrm{op}^2 C_w }{\alpha} \,,
\end{equation}
with probability at least $1 - 2\exp(-d)$.
By the optimality conditions of~\cref{eq:app:risk-def-duality} we have that $\hat{f}_\dataidx \in \partial_z \lossfun(y_\dataidx, z_\dataidx)$. 
To bound the norm of $\hat{\f}$, since $\lossfun$ is pseudo-Lipschitz of some order $k_L$ with constant $C''>0$, and $\frac{1}{\sqrt{n}}\hat{\z}$ is bounded with high probability, one can adopt the following
\begin{equation}
    \frac{1}{n} \|\hat{\f}\|_2^2 \leq \frac{C''}{n}\sum_\dataidx (1+|z_\dataidx|^{k_L-1} +|y_\dataidx|^{k_L-1})^2 \,,
\end{equation}

An application of the Borel–Cantelli lemma then ensures that the above bounds on $\hat{\w}, \hat{\z},\hat{\f}$ hold almost surely.
\end{proof}

\begin{remark}\label{rmk:app:absolute-val-regularization}
If we just suppose that $\regfun(w) \geq C_R \abs{w}$ for some $C_R>0$, by the equivalence of norms in any finite dimension, we have that $\frac{1}{d} \norm{\w}_2^2 \leq \norm{\w}_1 \leq C_w d$. This allows us to apply~\Cref{thm:gmt} in any finite dimension $d$ but the convergence of the extremizers of the high dimensional problem has to be considered separately.
\end{remark}

\begin{proof}[\Cref{thm:lower_b}]
The start of our analysis is the problem in \cref{eq:app:risk-def-duality} with the restriction of the extremization domain
\begin{equation}
    \minriskfun_d = 
    \inf_{\w \in \mathcal{S}_{C,d}, \z\in\mathcal{S}_{C,n}} 
    \sup_{\f \in \mathcal{S}_{C,n}}
    \frac{1}{d}
    \qty[ 
        \frac{1}{\sqrt{d}} \f^\top \datamat \w
        - \f^\top \z 
        + \sum_{\dataidx=1}^{n} \lossfun\qty(y_\dataidx, z_\dataidx)
        + \sum_{\dimidx=1}^{d} \regfun\qty(w_\dimidx) 
    ] \,,
\end{equation}
with $C$ the constant determined by~\Cref{prop:app:compactness-high-dim}.
Additionally we introduce the norm $q = \norm{\w}_2^2 / d$ and the overlap with the teacher $m = \wteacher^\top \w / d$
obtaining
\begin{align}
    \minriskfun_d = 
    \inf_{\w, \z, q, m} 
    \sup_{\f, \eta, \nu}
    \frac{1}{d}
    \Bigg[ &
        \frac{1}{\sqrt{d}} \f^\top \datamat \w
        - \f^\top \z 
        + \sum_{\dataidx=1}^{n} \lossfun\qty(y_\dataidx, z_\dataidx)
        + \sum_{\dimidx=1}^{d} \regfun\qty(w_\dimidx) + \nu (m d - \wteacher^\top \w)
        + \frac{\eta}{2} (\norm{\w}_2^2 - q d)
    \Bigg]
\end{align}
where because of Cauchy-Schwartz we have that $q \geq m^2/\rho$, where recall that by~\Cref{ass:teacher-channel} $\norm{\wteacher}_2^2 / d \to \rho$.

Now we want to separate the data dependence between the teacher subspace and the one orthogonal to the teacher subspace. 
We define $\s = \datamat \wteacher / \sqrt{d \rho}$ and introduce an independent Gaussian matrix $\Tilde{\datamat}$ where $\Tilde{X}_{ij} \sim \mathcal{N}(0,1)$ and we have that in distribution is the same as 
\begin{align}
    \minriskfun_d = 
    \inf_{\w, \z, m, q} 
    \sup_{\f, \nu, \eta} &
    \frac{m}{\sqrt{\rho}} \frac{\f^\top \s}{d} + 
    \frac{1}{\sqrt{d^3}} \f^\top \Tilde{\datamat} \qty(\w - \frac{m}{\rho}\wteacher)
    - \frac{\f^\top \z}{d}
    + \frac{1}{d} \sum_{\dataidx=1}^{n} \lossfun\qty(y_\dataidx, z_\dataidx) \\
    &+ \frac{1}{d} \sum_{\dimidx=1}^{d} \regfun\qty(w_\dimidx)
    + \nu m
    - \nu \frac{\wteacher^\top \w}{d} 
    + \frac{\eta}{2} \frac{\norm{\w}_2^2}{d}
    - \frac{\eta q}{2} \,.
\end{align}

Having expressed $\minriskfun_d$ in the above min-max form, we recognize a problem of the form of~\cref{eq:gmt:primary-problem} from~\Cref{thm:gmt} with 
$\mathbf{G} = \tilde{\datamat}$. 
We can can therefore write down a lower bound as
\begin{equation}
    \forall c \in \RR\,, \quad  \Pr[\minriskfun_d < c] \leq 2 \Pr[\tilde{A}_d < c] \,,
\end{equation}
where $\tilde{A}_d$ denotes the Auxiliary Optimization (A.O.) problem defined by~\cref{eq:gmt:auxiliary-problem} in~\Cref{thm:gmt}:
\begin{align}
    \minauxprob_d = 
    \inf_{\w, \z, m, q} 
    \sup_{\f, \nu, \eta} &
    \frac{m}{\sqrt{\rho}} \frac{\f^\top \s}{d} 
    - \frac{\norm{\f}_2}{\sqrt{d}} \frac{1}{d}\gaussvecone^\top \qty(\w - \frac{m}{\rho} \wteacher)
    + \frac{1}{\sqrt{d}} \norm{\w - \frac{m}{\rho} \wteacher}_2 \frac{\gaussvectwo^\top \f}{d}
    - \frac{\f^\top \z}{d} \\
    &+ \frac{1}{d} \sum_{\dataidx=1}^{n} \lossfun\qty(y_\dataidx, z_\dataidx) 
    + \frac{1}{d} \sum_{\dimidx=1}^{d} \regfun\qty(w_\dimidx)
    + \nu m
    - \nu \frac{\wteacher^\top \w}{d} 
    + \frac{\eta}{2} \frac{\norm{\w}_2^2}{d}
    - \frac{\eta q}{2} \,.
\end{align}
Given the two constraints introduced at the beginning we have the following
\begin{equation}
    \frac{1}{d} \norm{\w - \frac{m}{\rho} \wteacher}_2^2 = q - \frac{m^2}{\rho} \,.
\end{equation}
Now we optimize over the direction of $\f$ and define $\kappa = \norm{\f}_2 / \sqrt{d}$. Thus after the direction optimization we have
\begin{align}
    \minauxprob_d = 
    \inf_{\w, \z, m, q} 
    \sup_{\kappa, \nu, \eta} &
    \frac{\kappa}{\sqrt{d}} 
    \norm{\frac{m}{\sqrt{\rho}} \s + \sqrt{q - \frac{m^2}{\rho}} \gaussvectwo - \z}_2
    - \frac{\kappa}{d} \gaussvecone^\top \qty(\w - \frac{m}{\rho} \wteacher) \\
    &+ \frac{1}{d} \sum_{\dataidx=1}^{n} \lossfun\qty(y_\dataidx, z_\dataidx) 
    + \frac{1}{d} \sum_{\dimidx=1}^{d} \regfun\qty(w_\dimidx)
    + \nu m
    - \nu \frac{\wteacher^\top \w}{d} 
    + \frac{\eta}{2} \frac{\norm{\w}_2^2}{d}
    - \frac{\eta q}{2} \,.
\end{align}
Now we can invert $\min$ and $\sup$ and get an inequality. Additionally we use the fact that $\norm{\vv}_2 = \min _{\tau > 0} \frac{\tau}{2}+\frac{\norm{\vv}_2^2}{2 \tau}$ for any vector $\vv$ and get
\begin{align}
    \minauxprob_d \geq
    \sup_{\kappa, \nu, \eta}
    \inf_{\w, \z, m, q, \tau}  &
    \frac{\tau \kappa}{2}
    + \frac{\kappa}{2 \tau d} 
    \norm{\frac{m}{\sqrt{\rho}} \s + \sqrt{q - \frac{m^2}{\rho}} \gaussvectwo - \z}_2^2
    - \frac{\kappa}{d}\gaussvecone^\top \qty(\w - \frac{m}{\rho} \wteacher) \\
    &+ \frac{1}{d} \sum_{\dataidx=1}^{n} \lossfun\qty(y_\dataidx, z_\dataidx) 
    + \frac{1}{d} \sum_{\dimidx=1}^{d} \regfun\qty(w_\dimidx)
    + \nu m
    - \nu \frac{\wteacher^\top \w}{d} 
    + \frac{\eta}{2} \frac{\norm{\w}_2^2}{d}
    - \frac{\eta q}{2} \,.
\end{align}
We recognize the definition of Moreau envelope and get
\begin{align}
    \minauxprob_d \geq
    \sup_{\kappa, \nu, \eta}
    \inf_{\w, m, q, \tau}  &
    \frac{\tau \kappa}{2}
    + \frac{1}{d} \sum_{\dataidx=1}^{n} \frac{\kappa}{\tau}\mathcal{M}_{\frac{\tau}{\kappa} \lossfun(y_\dataidx, \cdot)} \qty(\frac{m}{\sqrt{\rho}} s_\dataidx + \sqrt{q - \frac{m^2}{\rho}} h_\dataidx) 
    + \frac{m}{\rho} \frac{\kappa}{d} \gaussvecone^\top \wteacher \\
    &+ \frac{1}{d} \sum_{\dimidx=1}^{d} \regfun\qty(w_\dimidx)
    - \frac{\kappa}{d}\gaussvecone^\top \w
    - \nu \frac{\wteacher^\top \w}{d} 
    + \frac{\eta}{2} \frac{\norm{\w}_2^2}{d}
    + \nu m
    - \frac{\eta q}{2} \,.
\end{align}
We now consider the last minimization over $\w$. To perform it we first complete the square 
\begin{align}
    \minauxprob_d \geq
    \sup_{\kappa, \nu, \eta}
    \inf_{\w, m, q, \tau}  &
    \frac{\tau \kappa}{2}
    + \frac{1}{d} \sum_{\dataidx=1}^{n} \frac{\kappa}{\tau}\mathcal{M}_{\frac{\tau}{\kappa} \lossfun(y_\dataidx, \cdot)} \qty(\frac{m}{\sqrt{\rho}} s_\dataidx + \sqrt{q - \frac{m^2}{\rho}} h_\dataidx) 
    + \frac{m}{\rho} \frac{\kappa}{d} \gaussvecone^\top \wteacher \\
    &+ \frac{1}{d} \sum_{\dimidx=1}^{d} \regfun\qty(w_\dimidx)
    + \frac{\eta}{2 d} \norm{\w - \frac{1}{\eta}(\kappa \gaussvecone + \nu \wteacher)}_2^2
    - \frac{1}{2 d \eta} \norm{\nu \wteacher + \kappa \gaussvecone}_2^2
    + \nu m
    - \frac{\eta q}{2} \,.
\end{align}
We now restrict the domain of $\eta$ to positive values, which is necessary for 
the completed square to define a minimization over $\w$. 
Since we are establishing a lower bound, this restriction of the supremum domain only preserves or strengthens the inequality. 
We then apply the definition of the Moreau envelope
\begin{align}
    \minauxprob_d \geq
    \sup_{\kappa, \nu, \eta}
    \inf_{m, q, \tau}  &
    \frac{\tau \kappa}{2}
    + \frac{1}{d} \sum_{\dataidx=1}^{n} \frac{\kappa}{\tau}\mathcal{M}_{\frac{\tau}{\kappa} \lossfun(y_\dataidx, \cdot)} \qty(\frac{m}{\sqrt{\rho}} s_\dataidx + \sqrt{q - \frac{m^2}{\rho}} h_\dataidx) 
    + \frac{m}{\rho} \frac{\kappa}{d} \gaussvecone^\top \wteacher \\
    &+ \frac{1}{d} \sum_{\dataidx = 1}^{d} \eta\mathcal{M}_{\frac{1}{\eta} \regfun(\cdot)} \qty(\frac{1}{\eta} \qty(\nu (\wteacher)_\dimidx + \kappa g_\dimidx) )
    - \frac{1}{2 d \eta} \norm{\nu \wteacher + \kappa \gaussvecone}_2^2
    + \nu m
    - \frac{\eta q}{2} \,,
\end{align}
where the optimization is taken over the set 
\begin{equation}\label{eq:app:faisable-set}
    \mathcal{S} = \qty{(m,q,\tau,\kappa, \nu, \eta) \in \RR^6 \mid \frac{m^2}{\rho} \leq q , a \leq q \leq b, \tau > 0, \eta > 0, \kappa \geq 0} \,.
\end{equation}

Define the following high-dimensional objective:
\begin{equation}
\begin{aligned}\label{eq:e-d-high-dim}
    \energyfun_d(m, q, \tau, \kappa, \nu, \eta) &\defeq 
    \frac{\tau \kappa}{2}
    + \frac{1}{d} \sum_{\dataidx=1}^{n} \frac{\kappa}{\tau}\mathcal{M}_{\frac{\tau}{\kappa} \lossfun(y_\dataidx, \cdot)} \qty(\frac{m}{\sqrt{\rho}} s_\dataidx + \sqrt{q - \frac{m^2}{\rho}} h_\dataidx) 
    + \frac{m}{\rho} \frac{\kappa}{d} \gaussvecone^\top \wteacher \\
    &+ \frac{1}{d} \sum_{\dataidx = 1}^{d} \eta\mathcal{M}_{\frac{1}{\eta} \regfun(\cdot)} \qty(\frac{1}{\eta} \qty(\nu (\wteacher)_\dimidx + \kappa g_\dimidx) )
    - \frac{1}{2 d \eta} \norm{\nu \wteacher + \kappa \gaussvecone}_2^2
    + \nu m
    - \frac{\eta q}{2} \,.
\end{aligned}
\end{equation} 

We are left with taking the high dimensional limit of $\energyfun_d(m, q, \tau, \kappa, \nu, \eta)$.
By~\Cref{lem:pseudo-lip-moreau}, the Moreau envelopes are pseudo-Lipschitz of finite order.
By the Strong Law of Large Numbers we have that 
\begin{equation}
    \frac{1}{d} \sum_{\dimidx=1}^d \eta\mathcal{M}_{\frac{1}{\eta} \regfun(\cdot)} \qty(\frac{1}{\eta} \qty(\nu (\wteacher)_\dimidx + \kappa g_\dimidx) ) 
    - \mathbb{E}\qty[
        \eta\mathcal{M}_{\frac{1}{\eta} \regfun(\cdot)} \qty(\frac{1}{\eta} \qty(\nu \tilde{w} + \kappa g) )
    ]
    \xrightarrow[]{a.s.} 0 \,,
\end{equation}
and likewise for the term with $\lossfun$.
Subsequently, by the Borel-Cantelli Lemma and standard concentration inequalities for Gaussian random variables, we obtain that, as $n,d\to \infty$, $\norm{\gaussvecone}_2^2/ d \xrightarrow[]{a.s.} 1$ and that for any $\vv \in \RR^d$, $\vv^\top \gaussvecone / d \xrightarrow[]{a.s.} 0$.

This yields the point-wise a.s. convergence of $\energyfun_d(m, q, \tau, \kappa, \nu, \eta)$ to the dimension-independent deterministic objective $\energyfun$:
\begin{equation}
     \energyfun_d(m, q, \tau, \kappa, \nu, \eta) \xrightarrow[d \rightarrow \infty]{a.s.} \energyfun(m, q, \tau, \kappa, \nu, \eta) \,,
\end{equation}
where we recall the definition of $\energyfun$ from~\cref{eq:low-dimensional-energy} as being
\begin{align}
    \energyfun(m, q, \tau, \kappa, \nu, \eta) &\defeq
    \frac{\tau \kappa}{2}
    + \alpha \mathbb{E}\qty[
        \frac{\kappa}{\tau}\mathcal{M}_{\frac{\tau}{\kappa} \lossfun(y, \cdot)} \qty(\frac{m}{\sqrt{\rho}} s + \sqrt{q - \frac{m^2}{\rho}} h) 
    ] \\
    &+ \mathbb{E}\qty[
        \eta\mathcal{M}_{\frac{1}{\eta} \regfun(\cdot)} \qty(\frac{1}{\eta} \qty(\nu \tilde{w} + \kappa g) )
    ]
    - \frac{1}{2 \eta} (\nu^2 \rho + \kappa^2 )
    + \nu m
    - \frac{\eta q}{2} \,.
\end{align}

We next translate the pointwise convergence of $\energyfun_d$ to $\energyfun$ into convergence of the extremal values associated with the sup-inf objective. This is done in~\Cref{lem:compact,lem:coerciv} and the proof is then completed.
\end{proof}

By~\Cref{ass:lossregfun} the expectations of Moreau operators and thus $\energyfun_d(m, q, \tau, \kappa, \nu, \eta)$ are equi-continuous in the parameters  $m, q, \tau, \kappa, \nu, \eta$. 
Therefore, by the Arzela-Ascoli Theorem, $\energyfun_d(m, q, \tau, \kappa, \nu, \eta)$ converges uniformly on compact sets. 
The following Lemma shows that the sequence of extremizers $m, q, \tau, \kappa, \nu, \eta$ can be restricted to fixed compact sets.

\begin{lemma}\label{lem:compact}
Under~\Cref{ass:lossregfun} there exists a compact set $\mathcal{S}$ with a diameter independent of $d$ such that the extremizers of $\mathcal{E}_d$ in~\cref{eq:e-d-high-dim} which we denote by 
$(\hat{m}_d, \hat{q}_d, \hat{\tau}_d, \hat{\kappa}_d, \hat{\nu}_d, \hat{\eta}_d)$
satisfy
\begin{equation}\label{eq:app:set-compact-diameter}
    \Pr[
        \cap_{d \geq d_0} 
        \qty{
            (\hat{m}_d, \hat{q}_d, \hat{\tau}_d, \hat{\kappa}_d, \hat{\nu}_d, \hat{\eta}_d)
            \in \mathcal{S}
        }
    ] \rightarrow 1 \,,
\end{equation}
as $d_0 \rightarrow \infty$.
\end{lemma}

\begin{proof}[\Cref{lem:compact}]
From~\Cref{prop:app:compactness-high-dim} we have that the optimal values of $\hat{q}_d$ and $\hat{\kappa}_d$ are finite because $q_d = \norm{\w}_2^2 / d$ and $\kappa_d = \norm{\f}_2 / d$. 
Moreover, the finiteness of the optimal value for $\hat{m}_d$ follows from an application of the Cauchy-Schwarz inequality. 

We want now to prove that the optimal value of the remaining parameters is finite as well.
We notice the following uniform upper bound of the Moreau envelope of the regularization
\begin{equation}\label{eq:app:upper-bound-moreau}
    \frac{1}{d} \eta\mathcal{M}_{\frac{1}{\eta} \regfun(\cdot)} \qty( 
    \frac{1}{\eta}\qty(\nu (\wteacher)_i + \kappa g_i) 
    ) \leq \frac{1}{d} \regfun(0) + \frac{1}{2 d \eta} \qty(\nu (\wteacher)_i + \kappa g_i)^2 \,.
\end{equation} 
By setting $m=0$ and taking $\tau \to 0^+$ in the inner problem, the above inequality implies that for any fixed $\kappa$ and $\nu$, as $\eta\to\infty$ the objective $\mathcal{E}_d$ tends to $-\infty$. 
This forces the optimal $\hat{\eta}_d$ to remain bounded.

Similarly, with $\eta_d$ and $\kappa_d$ bounded, one may apply again~\cref{eq:app:upper-bound-moreau} in the definition of $\mathcal{E}_d$ yields that the terms depending on $\nu$ satisfy
\begin{equation}
    \frac{1}{d} \sum_{\dataidx = 1}^{d} \eta\mathcal{M}_{\frac{1}{\eta} \regfun(\cdot)} \qty(\frac{1}{\eta} \qty(\nu (\wteacher)_\dimidx + \kappa g_\dimidx) )
    - \frac{1}{2 d \eta} \norm{\nu \wteacher + \kappa \gaussvecone}_2^2
    + \nu m 
    \leq 
    \regfun(0) + \nu m\,.
\end{equation}
For any fixed $\varepsilon > 0$, setting $m = \varepsilon$ in the inner minimization gives an upper bound on $\mathcal{E}_d$. As $\nu \to +\infty$, the term $\nu \varepsilon$ dominates, forcing $\mathcal{E}_d \to -\infty$. Similarly, setting $m = -\varepsilon$ yields $\mathcal{E}_d \to -\infty$ as $\nu \to -\infty$. Since the supremum of $\mathcal{E}_d$ over $\nu$ must be finite, the optimal $\hat{\nu}_d$ is necessarily bounded.

Finally, for the parameter $\tau$, note that for fixed and finite $\kappa$ we have
\begin{equation}
    \lim_{\tau \rightarrow +\infty} 
    \frac{1}{d} \frac{\kappa}{\tau}\mathcal{M}_{\frac{\tau}{\kappa} \lossfun(y_\dataidx, \cdot)} \qty(
        \frac{m}{\sqrt{\rho}} s_\dataidx + \sqrt{q - \frac{m^2}{\rho}} h_\dataidx 
    ) = 
    \frac{1}{d} \inf_{z \in \RR} \lossfun( y_\dataidx,z)
\end{equation}
and by \Cref{ass:lossregfun} the right-hand side is lower bounded. Hence, $\mathcal{E}_d(m,q,\tau,\kappa,\nu,\eta)$ tends to $+\infty$ as $\tau\to+\infty$, which implies that the optimal $\hat{\tau}_d$ must also be bounded.

We conclude that there exists an $R>0$, independent of $d$, such that
\begin{equation}
    \abs{\hat{m}_d} \leq R \,, \abs{\hat{q}_d} \leq R \,, \abs{\hat{\tau}_d}\leq R\, , \abs{\hat{\kappa}_d} \leq R \,, \abs{\hat{\nu}_d} \leq R \,, \abs{\hat{\eta}_d} \leq R\,.
\end{equation}
That is, the extremizers $(\hat{m}_d, \hat{q}_d, \hat{\tau}_d, \hat{\kappa}_d, \hat{\nu}_d, \hat{\eta}_d)$ lie in the compact set $\mathcal{S}\cap B(R)$, where $\mathcal{S}$ is defined in~\cref{eq:app:faisable-set}. 
This completes the proof.
\end{proof}

In light of \Cref{lem:compact}, we define $\tilde{\mathcal{E}}_d$ to be the restriction of $\mathcal{E}_d$ to the compact set $\mathcal{S}$ defined from the previous result.

\begin{lemma}\label{lem:coerciv}
    In the high-dimensional limit $d,n\to\infty$ of~\Cref{ass:high-dimensional-limit} we have that
    \begin{equation} 
        \abs{ \sup_{\kappa, \nu, \eta}
    \min_{m, q, \tau} 
        \tilde{\mathcal{E}}_d(m, q, \tau, \kappa, \nu, \eta) - \minriskfun} 
        \xrightarrow[d\to\infty]{a.s.} 0 
    \end{equation}
    where $\minriskfun$ is defined in~\cref{eq:min-risk-limit-def}.
\end{lemma}

\begin{proof}[\Cref{lem:coerciv}] 
Let $\mathcal{S}_{d,C}$ denote the following event
\begin{equation}
  \mathcal{S}_{d,C} \coloneqq \qty{  \frac{1}{d} \norm{\mathbf{s}}_1 < C, \frac{1}{d}\norm{\gaussvecone}_1 < C, \frac{1}{d}\norm{\mathbf{\tilde{w}}}_1 < C, \frac{1}{d}\norm{\mathbf{h}}_1 < C} \,,
\end{equation}
where $C$ denotes a fixed positive constant.
Then, concentration of  sub-Gaussian random variables \citep{vershynin2018high} and the Borel-Cantelli Lemma imply that $\exists C > 0$ such that:
\begin{equation}
    \lim_{d \rightarrow \infty}\Pr[ \mathcal{S}_{d,C}] = 1
\end{equation}
\Cref{ass:lossregfun} then implies that over $\mathcal{S}_{d,C}$, 
the objective $\mathcal{E}_d$ is equicontinuous in $m,q,\tau,\kappa,\nu,\eta$. The Lemma then follows by noting that point-wise convergence on a dense set for equicontinuous functions implies point-wise convergence everywhere.
\end{proof}

We are now interested in showing an equivalent description of the limiting energy that comes after the minimization over the low dimensional parameters $(m,q,\tau, \kappa, \nu, \eta)$.

\begin{corollary}[Stationary Condition]\label{cor:extr_se}
Any $(m, q, \tau, \kappa, \nu, \eta) \in \mathcal{S}^\star$ satisfies  the following set of equations
\begin{equation}\label{eq:app:self-consistent-eqs}
\begin{aligned}
    \partial_{\eta} :&\ 
    q =  
    \mathbb{E}\qty[
        \mathcal{P}_{\frac{1}{\eta} \regfun(\cdot)} \qty(\frac{1}{\eta} (\nu \tilde{w} + \kappa g))^2
    ] \\
    \partial_{\nu} :&\ m = \mathbb{E}\qty[
        \tilde{w} \mathcal{P}_{\frac{1}{\eta} \regfun(\cdot)} \qty(\frac{1}{\eta} (\nu \tilde{w} + \kappa g))
    ] \\
    \partial_{\kappa} :&\ \tau = \mathbb{E}\qty[
        g \mathcal{P}_{\frac{1}{\eta} \regfun(\cdot)} \qty(\frac{1}{\eta} (\nu \tilde{w} + \kappa g))
    ] \\
    \partial_{m} :&\ \nu = \alpha \mathbb{E}\qty[
        \qty(\frac{s}{\sqrt{\rho}} - \frac{\sfrac{m}{\rho} h}{\sqrt{q - \frac{m^2}{\rho}}}) \mathcal{P}_{\frac{\tau}{\kappa}\lossfun(y,\cdot)} \qty(\frac{m}{\sqrt{\rho}} s + \sqrt{q  - \frac{m^2}{\rho^2}} h)
    ] \\
    \partial_{\tau} :&\ \kappa^2 = 
    \frac{\alpha \kappa^2}{\tau^2} \mathbb{E}\qty[
        \qty(
            \frac{m}{\sqrt{\rho}} s + \sqrt{q  - \frac{m^2}{\rho}} h - 
            \mathcal{P}_{\frac{\tau}{\kappa}\lossfun(y,\cdot)} \qty(\frac{m}{\sqrt{\rho}} s + \sqrt{q  - \frac{m^2}{\rho}} h)
        )^2
    ] \\
    \partial_{q} :&\ \alpha \frac{\kappa}{\tau} - \eta = \frac{\alpha}{\sqrt{q  - \frac{m^2}{\rho}}} 
    \mathbb{E}\qty[
        h \, \mathcal{P}_{\frac{\tau}{\kappa}\lossfun(y,\cdot)} \qty(\frac{m}{\sqrt{\rho}} s + \sqrt{q  - \frac{m^2}{\rho}} h)
    ]
\end{aligned}
\end{equation}
where the expectation is taken over $s,g,h \sim \mathcal{N}(0,1)$, $\tilde{w} \sim \pi_{\wteacher}$ and $y \sim P(\cdot \mid s)$ from~\Cref{ass:teacher-channel}.
\end{corollary}

\begin{proof}[\Cref{cor:extr_se}] 
These fixed point equations can be found by application of the two following rules
\begin{equation}
    \frac{\partial \mathcal{M}_{\tau f}(z)}{\partial z}=z-\mathcal{P}_{\tau f}(z) \,, \quad
    \frac{\partial \left(\frac{1}{\tau}\mathcal{M}_{\tau f}(z)\right)}{\partial \tau}=-\frac{\left(z-\mathcal{P}_{\tau f}(z)\right)^2}{2 \tau^2} \,.
\end{equation}
they hold almost everywhere because of the regularity assumptions made, specifically~\Cref{ass:lossregfun}. Non-emptiness of proximals is because of assumptions made on $\lossfun$ and $\regfun$ as described in~\Cref{sec:app:from-weak-convex-to-lipshitz}.
\end{proof}

\begin{remark}
    Since $\mathcal{M}$ is pseudo-Lipschitz from \Cref{ass:lossregfun} and \Cref{lem:pseudo-lip-moreau}, it is differentiable almost everywhere. Dominated convergence theorem (finite differences dominated by the pseudo-Lipschitzness) then ensures that the above expressions are well-defined.
\end{remark}

\begin{corollary}\label{thm:moreau-form-energy}
For any $(m^\star, q^\star, \tau^\star, \kappa^\star, \nu^\star, \eta^\star) \in \mathcal{S}^\star$ defined in~\cref{eq:def-s-star}, it holds that
\begin{equation}\label{eq:simplified-optimal-value}
    \minriskfun = \alpha \mathbb{E}\qty[
        \lossfun\qty(y, \mathcal{P}_{\frac{\tau^\star}{\kappa^\star}\lossfun(y,\cdot)} \qty(\frac{m^\star}{\sqrt{\rho}} s + \sqrt{q^\star  - \frac{(m^\star)^2}{\rho}} h))
    ] + \mathbb{E}\qty[
        \regfun\qty(\mathcal{P}_{\frac{1}{\eta^\star} \regfun(\cdot)} \qty(\frac{1}{\eta^\star} (\nu^\star \tilde{w} + \kappa^\star g)))
    ] \,,
\end{equation}
where the expectation is taken over $s,g,h \sim \mathcal{N}(0,1)$, $\tilde{w} \sim \pi_{\wteacher}$ and $y \sim P(\cdot \mid s)$ from~\Cref{ass:teacher-channel}.
\end{corollary}

\begin{proof}[\Cref{thm:moreau-form-energy}]
First we notice that one can always write the Moreau envelope as
\begin{equation}
    \mathcal{M}_{f} (\omega) = f\qty(z) + \frac{1}{2} \qty(z - \omega)^2 \,,
\end{equation}
for any $z \in \mathcal{P}_{f} (\omega)$, which is non-empty by~\Cref{ass:lossregfun,ass:weak_convexity} about $\lossfun$ and $\regfun$.

We can thus exchange the two Moreau envelopes in~\cref{eq:low-dimensional-energy} with the previous identities.
Then it follows from algebraic manipulation of the equations presented in \Cref{cor:extr_se} the final form we are interested in.
\end{proof}

\section{Proofs for the Upper Bound}\label{sec:app:upper-bound}

We will now construct a sequence of iterates of the approximate message passing algorithm attaining the lower bound defined by~\Cref{thm:lower_b} with the goal in mind of proving~\Cref{lem:stab}.

We will analyze two different instances of~\Cref{algo:gamp} with different choices of non linear functions $\{\inden[t]\}_{t\geq 0}$ and same initialization given by~\cref{eq:init-gamp} and same choice of $\{\outden[t]\}_{t\geq 0}$. In both cases we chose $\outden[t] = \outden[\star]$ acting component wise.

We will refer to the first algorithm as the \textit{unprojected algorithm} and have $\forall t \geq 0$ the non linear function to be
\begin{equation}\label{eq:app:f}
    \inden[t](\w_i) = \inden[\star](\w_i) \defeq \argmin_{z \in \mathbb{R}} \qty[ \regfun(z) + \frac{\eta^\star}{2} (z - \w_i)^2 ] \,, 
\end{equation}
where $\inden[\star]$ comes from~\cref{eq:output-input-denoisers-amp} acting component wise.
The second algorithm will be the \textit{projected version} of it where $\forall t \geq 0$ the non linear function is chosen as
\begin{equation}\label{eq:app:f-c}
    \tilde{\inden[t]}(\w, c) = \frac{c}{\norm{\inden[t](\w)}_2}\inden[t](\w) \,,
\end{equation}
with the definition of $\inden[t]$ from~\cref{eq:app:f}.

Additionally we define the pre-compued value of the rescaled norm as
\begin{equation}\label{eq:norm_const}
    c \coloneqq \frac{\nu^\star}{\eta^\star} + \frac{\kappa^\star}{\eta^\star} \,.
\end{equation}

Our choice of initialization (\cref{eq:init-gamp}) and Theorem \ref{lem:state-evolution-general-covariates} imply that the \textit{uprojected} algorithm for any $t \in \mathbb{N}$, the state-evolution parameters remain at fixed values:

\begin{proposition}\label{prop:fixdpt}
For any  $(m^\star, q^\star, \tau^\star, \kappa^\star, \nu^\star, \eta^\star) \in \mathcal{S}^\star$, let $c^\star$ be as defined in~\cref{eq:norm_const} and let $\w^0$ be initialized as~\cref{eq:init-gamp}.
Consider~\Cref{algo:gamp} with $\outden[t]=\outden[\star]$ and $\inden[t]=\inden(\w)$ from~\cref{eq:app:f} (the \textit{unprojected} algorithm).
Then, the parameters $\beta, \Omega, \mu, \Sigma$ defined as
\begin{equation}\label{eq:app:map-se-gordon}
    \beta = \frac{m^\star}{\rho} \,, \qquad 
    \Omega = q^\star - \frac{(m^\star)^2}{\rho} \,, \qquad 
    \mu = \frac{\nu^\star}{\eta^\star} \,, \qquad 
    \Sigma = \frac{\kappa^\star}{\eta^\star} \,,
\end{equation}
satisfy the fixed point conditions for the state-evolution equations in \cref{eq:beta-def,eq:omega-def,eq:mu-def,eq:sigma-def}. 
Furthermore the estimate $\hat{\w}^t$ for any $t \in \mathbb{N}$:
\begin{equation}\label{eq:prop:norm-constant}
    \frac{1}{d} \norm{\hat{\w}^t}^2 \xrightarrow[d \rightarrow \infty]{a.s.} c \,.
\end{equation}
\end{proposition}
\begin{proof}[\Cref{prop:fixdpt}]
With the choice of initialization as in \cref{eq:init-gamp}, the fixed point conditions for the state evolution 
\cref{eq:mu-def,eq:sigma-def} under our choice of $\outden[t] = \outden[\star]$ become:
\begin{align}
    \mu &= \alpha \mathbb{E}\qty[
        v^\star \mathcal{P}_{\frac{\tau^\star}{\kappa^\star}\lossfun(y,\cdot)} \qty(\frac{m^\star}{\sqrt{\rho}} v^\star + v )
    ] \,, 
    \label{eq:app:se-mu-specified} \\
    \Sigma &= \alpha \mathbb{E}\qty[
        \mathcal{P}_{\frac{\tau^\star}{\kappa^\star}\lossfun(y,\cdot)} \qty(\frac{m^\star}{\sqrt{\rho}} v^\star + v )^2
    ] \,,
    \label{eq:app:se-sigma-specified}
\end{align}
where $v^\star \sim \mathcal{N}(0,1)$ and $v \sim \mathcal{N}(0,q^\star - \sfrac{(m^\star)^2}{\rho})$.
We next establish that \cref{eq:app:se-sigma-specified,eq:app:se-mu-specified} follow from the fourth and fifth stationary conditions in \cref{eq:app:self-consistent-eqs}. To show this, we follow the argument presented in \citep[Appendix~C.2]{loureiro2021learning}. The computation proceeds in two steps: first, we substitute the expressions for $\mu$ and $\Sigma$ from \cref{eq:app:map-se-gordon} into the relevant conditions in \cref{eq:app:self-consistent-eqs}; second, we apply a change of variables followed by Gaussian integration to demonstrate the equivalence.

\Cref{eq:beta-def,eq:omega-def} become in our case
\begin{align}
    \beta &= \frac{1}{\rho} \mathbb{E}\qty[ \tilde{w} \inden\qty(\mu \tilde{w} + u, c) ] \,, 
    \label{eq:app:se-beta-specified} \\
    \Omega &= \mathbb{E}\qty[ \inden\qty(\mu \tilde{w} + u, c)^2 ] - \frac{(m^\star)^2}{\rho} \,, 
    \label{eq:app:se-omega-specified}
\end{align}
where, under~\Cref{ass:teacher-channel}, we have $\rho = \mathbb{E}\qty[(\tilde{w})^2]$ and $u \sim \mathcal{N}(0, \sfrac{(\kappa^\star)^2}{(\eta^\star)^2})$. 
By substitution of the value of $\beta$ and $\Omega$ in \cref{eq:app:map-se-gordon} we directly see that \cref{eq:app:se-beta-specified} matches the second equation from~\Cref{cor:extr_se} and that~\cref{eq:app:se-omega-specified} matches the first one.

Similarly, by inductively applying the above fixed-point iterations and Theorem \ref{lem:state-evolution-general-covariates}, we obtain that the value of  $\mathbb{E}[\sfrac{1}{d}\|\w\|_2^2]$, remains constant across iterations and equal to the value $c$ defined in~\cref{eq:norm_const}.
\end{proof}

\begin{proposition}\label{prop:algo-2-norm}
Let $\w^t$ be the iterate of the unprojected algorithm with initialization given by~\cref{eq:init-gamp} and with the choice of $\inden[t]$ being proximal as per~\cref{eq:app:f}.
Let $\bar{\w}^t$ denote the iterates of the projected algorithm 
with initialization given by~\cref{eq:init-gamp} and with the choice of
$\inden[t](\w)=\tilde{\inden}(\w,c)$ defined in~\cref{eq:app:f-c} where the constant $c>0$ is defined above in~\cref{eq:norm_const}.
Then, for all $t \in \mathbb{N}$
\begin{equation}
    \frac{1}{d} 
    \norm{ \w^{t}  - \bar{\w}^t }^2 
    \xrightarrow[d\to\infty]{a.s.} 
    0 \,.
\end{equation}
\end{proposition}

\begin{proof}[\Cref{prop:algo-2-norm}]
We proceed by induction. Let $\bar{\p}^t$ denote the pre-activations corresponding to Algorithm \ref{algo:gamp} with iterates replaced by $\bar{\w}^t$ defined above, i.e.:
\begin{equation}
    \bar{\p}^t = \frac{1}{\sqrt{d}}
    \datamat \inden[t] \qty( \bar{\w}^t ) - b^t
    \outden[t-1] \qty(\bar{\p}^{t-1}, \y)\
\end{equation}

Suppose that the following claim holds at time $t-1$:
\begin{equation}
    \frac{1}{d}\norm{\w^{t-1}-\bar{\w}^{t-1}}^2 \xrightarrow[d\to\infty]{a.s.} 0 \,, \quad \frac{1}{d}\norm{\p^{t-1}-\bar{\p}^{t-1}}^2 \xrightarrow[d\to\infty]{a.s.} 0 \,. \tag{$\star$}\label{eq:induction-hyp}
\end{equation}

We have, by line $5$ in~\Cref{algo:gamp}:
\begin{align}
    \frac{1}{d}\norm{\w^{t}-\bar{\w}^t}^2 &= \frac{1}{d} \norm{\datamat}^2_{\mathrm{op}} \norm{g^t(\p^t, \y)-g^t(\bar{\p}^t, \y)}^2_2 + \frac{1}{d}\norm{
        \inden(\w^{t-1})
        - 
        \tilde{\inden}(\bar{\w}^{t-1},c)
    }^2 \\
    &\leq \frac{1}{d} \norm{\datamat}^2_{\mathrm{op}} \norm{g^t(\p^t, \y)-g^t(\bar{\p}^t, \y)}^2_2
    + \frac{1}{d}\norm{
        \tilde{\inden}(\w^{t-1},c)
        - 
        \tilde{\inden}(\bar{\w}^{t-1},c)
    }^2\\
    &\quad +\frac{1}{d}\norm{
        \inden\qty(\w^{t-1})
        - 
        \tilde{\inden}(\w^{t-1},c)
    }^2 \,.
\end{align}

We recall from~\cref{eq:bound-operator-norm} that $\frac{1}{\sqrt{d}}\norm{\datamat}_{\mathrm{op}} \leq C_{\text{op}}$, $a.s$ as $d \rightarrow \infty$ for some constant $C_{\text{op}}>0$. 
Hence, the first two-terms vanish by the induction hypothesis \eqref{eq:induction-hyp} and pseudo-lipschitzness of $g$ while the last two terms vanishes by the pseudo-lipschitzness of $\tilde{\inden}$ and \cref{eq:prop:norm-constant}.
\end{proof}

\vspace{1em}
We can now proceed to show that~\Cref{ass:stab} implies the convergence of the message passing algorithm.
\begin{proof}[\Cref{lem:stab}]
In the following proof, we will denote 
\begin{equation}
    \Omega = q^{\star}-\frac{(m^{\star})^{2}}{\rho} \quad \mbox{and} \quad \Sigma = \frac{\kappa^{\star}}{\eta^{\star}}.
\end{equation}
We will also use the following shorthands for the update functions of our AMP iteration, for any real valued input $x$:
\begin{equation}
    f(x) = f^{\star}\left(\frac{\nu^{*}}{\eta^{*}}\tilde{w}+x\right) \,, \quad
    g(x) = g^{\star}\left(\frac{m^{*}}{\rho}s+x,y\right) \,,
\end{equation}
where $\tilde{w}$ and $s$ are real-valued random variables as defined in Definition \ref{def:replicon-stability}.
We are interested in the evolution of the large system limits of the quantities 
\begin{align}
    \frac{1}{d}\norm{\mathbf{p}^{t}-\mathbf{p}^{t-1}}_{2}^{2} &= \frac{1}{d}\norm{\mathbf{p}^{t}}_{2}^{2}+\frac{1}{d}\norm{\mathbf{p}^{t-1}}_{2}^{2}-2\frac{1}{d}\langle \mathbf{p}^{t},\mathbf{p}^{t-1} \rangle, \\
    \frac{1}{d}\norm{\mathbf{w}^{t}-\mathbf{w}^{t-1}}_{2}^{2} &= \frac{1}{d}\norm{\mathbf{w}^{t}}_{2}^{2}+\frac{1}{d}\norm{\mathbf{w}^{t-1}}_{2}^{2}-2\frac{1}{d}\langle \mathbf{w}^{t},\mathbf{w}^{t-1} \rangle.
\end{align}
Owing to~\Cref{lem:state-evolution-general-covariates} applied to the \textit{unprojected} algorithm we get~\Cref{prop:algo-2-norm}, then because of~\Cref{prop:fixdpt} it holds that, for any $t \geq 0$, 
\begin{equation}
    \lim_{d \to \infty} \frac{1}{d}\norm{\mathbf{p}^{t}}_{2}^{2} = \Omega \quad \mbox{and} \quad  \lim_{d \to \infty} \frac{1}{d}\norm{\mathbf{w}^{t}}_{2}^{2} = \Sigma;
\end{equation}
furthermore, the quantities 
\begin{equation}
    \lim_{d \to \infty} \frac{1}{d}\langle \mathbf{p}^{t},\mathbf{p}^{t-1} \rangle \quad \mbox{and} \quad \lim_{d \to \infty} \frac{1}{d}\langle \mathbf{w}^{t},\mathbf{w}^{t-1} \rangle,
\end{equation}
are well defined. From now on, we will respectively denote these quantities $\hat{C}_{t}$ and $C_{t}$, so that 
\begin{align}
    \lim_{d \to \infty} \frac{1}{d}\norm{\mathbf{p}^{t}-\mathbf{p}^{t-1}}_{2}^{2} &= 2(\Omega-\hat{C}_{t}), \\
    \lim_{d \to \infty} \frac{1}{d}\norm{\mathbf{w}^{t}-\mathbf{w}^{t-1}}_{2}^{2} &= 2(\Sigma-C_{t}).
\end{align}
Note that the Cauchy-Schwarz inequality ensures that $\hat{C}_{t}\leq \Omega$ and $C_{t} \leq \Sigma$.
To establish convergence of the AMP sequence, we thus need to study the convergence of the sequences $\{C_{t}\}_{t \geq 0}, \{\hat{C}_{t}\}_{t \geq 0}$. Using~\Cref{lem:state-evolution-general-covariates} and a similar argument to \cite{bolthausen2014iterative}, we may represent the state evolution equations governing the evolution of $C_{t},\hat{C}_{t}$ as follows  
\begin{align}
    C^{t+1} &= \mathbb{E}\left[g(\sqrt{\hat{C}_{t}}z_{1}+\sqrt{\Omega-\hat{C}_{t}}z'_{1})g(\sqrt{\hat{C}_{t}}z_{1}+\sqrt{\Omega-\hat{C}_{t}}z''_{1})\right], \\
    \hat{C}_{t} &= \alpha \mathbb{E}\left[f(\sqrt{C_{t}}z_{2}+\sqrt{\Sigma-C_{t}}z'_{2})f(\sqrt{C_{t}}z_{2}+\sqrt{\Sigma-C_{t}}z''_{2})\right].
\end{align}
Starting with the first equation, we compute the partial derivative w.r.t. $\hat{C}_{t}$, so that we obtain 
\begin{align}
    \partial_{\hat{C}_{t}}C_{t+1} &= \frac{1}{\sqrt{\hat{C}_{t}}}\mathbb{E}\left[z_{1}g'(\sqrt{\hat{C}_{t}}z_{1}+\sqrt{\Omega-\hat{C}_{t}}z'_{1})g(\sqrt{\hat{C}_{t}}z_{1}+\sqrt{\Omega-\hat{C}_{t}}z''_{1})\right] \\
    &-\frac{1}{\sqrt{\Omega-\hat{C}_{t}}}\mathbb{E}\left[z'_{1}g'(\sqrt{\hat{C}_{t}}z_{1}+\sqrt{\Omega-\hat{C}_{t}}z'_{1})g(\sqrt{\hat{C}_{t}}z_{1}+\sqrt{\Omega-\hat{C}_{t}}z''_{1})\right].
\end{align}
Gaussian integration by parts (Stein's lemma) then yields 
\begin{equation}
    \partial_{\hat{C}_{t}}C_{t+1} = \mathbb{E}\left[g'(\sqrt{\hat{C}_{t}}z_{1}+\sqrt{\Omega-\hat{C}_{t}}z'_{1})g'(\sqrt{\hat{C}_{t}}z_{1}+\sqrt{\Omega-\hat{C}_{t}}z''_{1})\right].
\end{equation}
Integrating our the independent variables $z_{1}',z_{1}''$, we obtain that the above quantity is strictly positive so that, for any $t \geq0$, the function $C_{t+1}(.)$ is strictly increasing. Taking an additional derivative, we obtain 
\begin{align}
    \partial^{2}_{\hat{C}_{t}}C_{t+1} = \mathbb{E}\left[g'(\sqrt{\hat{C}_{t}}z_{1}+\sqrt{\Omega-\hat{C}_{t}}z'_{1})g'(\sqrt{\hat{C}_{t}}z_{1}+\sqrt{\Omega-\hat{C}_{t}}z''_{1})\right],
\end{align}
which is also positive, so that $C_{t+1}(.)$ is convex on $[0,\Omega]$. We can therefore bound both $C_{t+1}$ and its derivative with their values at $\hat{C}_{t} = \Omega$, giving the following upper bound
\begin{equation}
    \vert \partial_{\hat{C}_{t}}C_{t+1} \vert \leq \mathbb{E}\left[g'\left(\sqrt{\hat{C}_{t}}z_{1}\right)^{2}\right].
\end{equation}
A similar computation gives the upper bound 
\begin{equation}
    \vert \partial_{C_{t}}\hat{C}_{t} \vert \leq \alpha\mathbb{E}\left[f'\left(\sqrt{C_{t}}z_{2}\right)^{2}\right].
\end{equation}
Under the replicon condition \ref{def:replicon-stability}, we thus find that the iterations $\hat{C}_{t+1} = \mathcal{F}_{1}(\hat{C}_{t})$ and $C_{t+1} = \mathcal{F}_{2}(C_{t})$ are non-expansive, so that the only fixed point is given by $C = \Omega$ and $\hat{C} = \Sigma$, leading to 
\begin{align}
    \lim_{t \to \infty} \lim_{d \to \infty} \frac{1}{d}\norm{\mathbf{w}^{t}-\mathbf{w}^{t-1}}_{2}^{2} = 0 \quad \mbox{and} \quad \lim_{t \to \infty} \lim_{d \to \infty} \frac{1}{d}\norm{\mathbf{p}^{t}-\mathbf{p}^{t-1}}_{2}^{2} = 0.
\end{align}
Finally, assuming that the replicon stability condition in Definition \ref{def:replicon-stability} holds strictly gives the linear convergence of the AMP sequence and concludes this proof.
\end{proof}

\section{Proofs for Matching Upper and Lower Bounds}\label{sec:app:matching}

This section is dedicated to the proof of~\Cref{prop:physicality-fixed-point}.
In this section we will use the following auxiliary objective
\begin{equation}\label{eq:intermediate-amp-risk}
    \mathcal{C}_d(\p^t, \w^t) \defeq \frac{\alpha}{n} \sum_{\dataidx = 1}^n \lossfun\qty(
        y_\dataidx, 
        \tilde{
        \mathcal{P}
        }
        _{\frac{\tau^\star}{\kappa^\star} \lossfun(y_\dataidx, \cdot)} (p_\dataidx)
    )
    + \frac{1}{d} \sum_{\dimidx = 1}^d \regfun\qty(
        \tilde{
        \mathcal{P}
        }
        _{\frac{1}{\eta^\star}\regfun(\cdot)}(w_\dimidx)
    ) \,,
\end{equation}

\begin{lemma}\label{lem:error_conv}
    We define the risk of the iterates of~\Cref{algo:gamp} at time $t$ as begin $\riskfun_d(\what^t)$ where $\what^t = \inden[\star](\w^t)$.
    Under \Cref{ass:weak_convexity,ass:stab}, there exist constants $K> 0$ and $\rho^\prime <1$ such that the following holds for any $t$
    \begin{equation}\label{eq:convergence-energy}
        \abs{\riskfun_d(\what^t) - \mathcal{C}_d(\p^t, \w^t)} \leq  K (\rho^\prime)^t,
    \end{equation}
    with probability $\rightarrow 1$ as $n,d \rightarrow \infty$.
\end{lemma}

\begin{proof}[\Cref{lem:error_conv}]
From \Cref{algo:gamp}, the preactivations $\z^t$ at time $t$ can be expressed as
\begin{equation}
    \z^t \defeq \frac{1}{\sqrt{d}} \datamat f(\w^t) = \p^t + b^tg^{t-1}(\p^{t-1},\y),
\end{equation}
where $b^t$ is a scalar. 
The objective value at time $t$ is therefore given by
\begin{align}
    \riskfun_d(\what^t)
    &= \frac{\alpha}{n} \sum_{\dataidx = 1}^{n} \lossfun(y_\dataidx, z^t_\dataidx) 
    + \frac{1}{d} \sum_{\dimidx = 1}^{d} \regfun(\inden[\star](w^t_\dimidx))  \\
    &= \frac{\alpha}{n} \sum_{\dataidx = 1}^{n} \lossfun(y^t_\dataidx, p^t_\dataidx + b^t g^{t-1}(p^{t-1}_\dataidx, y_\dataidx)) 
    + \frac{1}{d} \sum_{\dimidx = 1}^{d} \regfun(\inden[\star](w^t_\dimidx)) \,.
\end{align}

The above objective depends on the correlation between $\p^t, \p^{t-1}$. 
We need to show that as $\frac{1}{n}\norm{\p^t-\p^{t-1}}^2_2 \rightarrow 0$, \textit{i.e.} the algorithm is converging $\riskfun_d(\what^t) \rightarrow \mathcal{C}_d(\p^t, \w^t)$.
From the specific choice of \(\outden[\star]\) from~\cref{eq:output-input-denoisers-amp}, we have that
\begin{equation}
    b^t \outden[\star](p^{t-1}_\dataidx, y_\dataidx) \xrightarrow[d \to \infty]{a.s.} \mathcal{P}_{\frac{\tau^\star}{\kappa^\star} \lossfun(y_\dataidx, \cdot)}(p_\dataidx^{t-1}) 
    - p^{t-1}_\dataidx \,,
\end{equation}
and thus
\begin{equation}
    \riskfun_d(\what^t) = \frac{\alpha}{n} \sum_{\dataidx=1}^n \lossfun\qty(
        y_\dataidx, 
        p_\dataidx^t 
        - p_\dataidx^{t-1} 
        + \tilde{\mathcal{P}}_{\frac{\tau^\star}{\kappa^\star} \lossfun(y_\dataidx, \cdot)}(p_\dataidx^{t-1})
    )
    + \frac{1}{d} \sum_{\dimidx = 1}^{d} 
    \regfun\qty(\inden[\star](w^t_\dimidx)) \,.
\end{equation}

Now, since $\lossfun$ is pseudo-Lipschitz of finite-order $k_L$ because of~\Cref{ass:lossregfun}, there exists a constant $L > 0$, such that:
\begin{align}
    \lossfun(y_\dataidx, p_\dataidx^t - p_\dataidx^{t-1} + \tilde{\mathcal{P}}_{\tau \lossfun(y, \cdot)}(p_\dataidx^{t-1})) &\leq 
    \lossfun(y_\dataidx, \tilde{\mathcal{P}}_{\tau \lossfun(y, \cdot)}(p_\dataidx^{t-1})) \\
    &+
    L \abs{p^t_\dataidx - p^{t-1}_\dataidx} 
    \qty(
        1 
        + \abs{\tilde{\mathcal{P}}_{\tau \lossfun(y, \cdot)}(p_\dataidx^{t-1})}^{k_L} 
        + \qty(p^t_\dataidx - p^{t-1}_\dataidx)^{k_L}
    ) \,.
\end{align}

Combining the previous equation in the definition of the risk and the definition of $\inden[\star]$ from~\cref{eq:output-input-denoisers-amp}, we have that
\begin{align}
    \abs{\riskfun_d(\what^t) - \mathcal{C}_d(\p^t, \w^t)} \leq 
    \underbrace{
        L \frac{\alpha}{n}\sum_{\dataidx=1}^n 
        \abs{p^t_\dataidx - p^{t-1}_\dataidx} 
        \qty(
            1
            + \abs{\tilde{\mathcal{P}}_{\tau \lossfun(y, \cdot)}(p_\dataidx^{t-1})}^{k_L} 
            + \abs{p^t_\dataidx - p^{t-1}_\dataidx}^{k_L}
        )
    }_{\delta}\,.
\end{align}

It suffices to show that $\delta \rightarrow 0$ as $t \rightarrow \infty$. 
We obtain by Cauchy-Schwarz inequality the following

\begin{equation}
    \delta \leq 
    \frac{1}{\sqrt{n}} \norm{\p^t - \p^{t-1}}_2
    \sqrt{
    \frac{1}{n} \sum_{\dataidx=1}^{n}
    \qty(
        1
        + \abs{\tilde{\mathcal{P}}_{\tau \lossfun(y, \cdot)}(p_\dataidx^{t-1})}^{k_L} 
        + \abs{p^t_\dataidx - p^{t-1}_\dataidx}^{k_L}
    )^2
    }\,.
\end{equation}

The convergence of $\frac{1}{\sqrt{n}}\norm{\p^t - \p^{t-1}}_2$ is established by~\Cref{lem:stab}. 
The remaining term can be upper bounded in probability by a constant as it is a pseudo-Lipschitz function by~\Cref{lem:state-evolution-general-covariates}.

Therefore, $\delta$ decays in $t$ with high probability in $n$, yielding~\cref{eq:convergence-energy} almost surely and completing the proof.
\end{proof}

\begin{lemma}\label{lem:aux_error}
From the definition of the auxiliary objective in~\cref{eq:intermediate-amp-risk} we have that for any $t \geq 1$:
\begin{equation}
     \mathcal{C}_d(\p^t, \w^t) \xrightarrow[n,d \to \infty]{a.s.} \minriskfun \,.
\end{equation}
\end{lemma}

\begin{proof}[\Cref{lem:aux_error}]
The function we are interested in is separable and thus amenable for~\Cref{lem:state-evolution-general-covariates}.
At fixed point, by the state-evolution description, we have that the distribution for 
$\p$ and $\w$ are
\begin{equation}
    \p \approx \frac{m^\star}{\rho} \s + \sqrt{q^\star - \frac{(m^\star)^2}{\rho}} \gaussvectwo \,, \quad \w \approx \frac{nu^\star}{\eta^\star} \wteacher + \frac{\kappa^\star}{\eta^\star} \gaussvecone \,,
\end{equation}
where $\mathbf{s}, \gaussvectwo \sim \mathcal{N}(\mathbf{0}, \mathrm{Id}_n)$ and $\gaussvecone \sim \mathcal{N}(\mathbf{0}, \mathrm{Id}_d)$ and $\wteacher \sim \pi_{\wteacher}$.
We can thus see by substituting that we obtain the specific form for the follows by~\Cref{thm:moreau-form-energy}. 
\end{proof}

\vspace{1em}
The proof of~\Cref{prop:physicality-fixed-point} follows from combining~\Cref{lem:aux_error,lem:error_conv}. 
The key observation is that both the \textit{projected} or the \textit{unprojected} algorithm converge to the same energy value.  
To see this, note that the projection onto the spherical shell amounts to rescaling by a factor that converges almost surely to a constant by~\Cref{prop:fixdpt} this implies 
convergences of all higher moments are converging for both algorithms iterates.
By the pseudo-Lipschitz of both $\lossfun$ and $\regfun$ from~\Cref{ass:lossregfun}, the auxiliary objective $\mathcal{C}_d(\p^t, \w^t)$ for both algorithms converges almost surely to the same limit characterized in~\Cref{thm:moreau-form-energy}, completing the argument.

\subsection{Proof of the Main Statement}

We now proceed to the proof of~\Cref{thm:main} under~\Cref{ass:weak_convexity,ass:stab,ass:lossregfun,ass:high-dimensional-limit,ass:teacher-channel}.
To establish the convergence of the statistic $\psi$ from~\Cref{thm:main} we leverage the following theorem which is a consequence of \citep[Theorem~24.5]{rockafellar1970convex}.

\begin{theorem}
\label{thm:conv-derivative}
Let $f_1, f_2 \cdots$ be a sequence of convex functions on $\RR$ converging almost-surely pointwise to $f$ on an open interval $O \subset \RR$. Then, for any $\varepsilon > 0$:
\begin{equation}\label{eq:grad_inc}
    \Pr[\cap_{d >d_0} 
        \qty{\partial f_d(s) \in \partial f(s) + \varepsilon B}
    ] \rightarrow 1,
\end{equation}
as $d_0 \rightarrow \infty$, and
where $B$ denotes the unit ball in $\RR$ and we indicate by $\partial f(s)$ the subdifferential of $f$ at the point $s$. 
\end{theorem}

\begin{proof}[\Cref{thm:conv-derivative}]
Let $\partial_{+} f_d(s), \partial_{-} f_d(s)$ denote the right and left derivatives of $f_d$ respectively defined as
\begin{equation}
    \partial_{+} f_d(s) = \lim_{t \downarrow 0^+} \frac{f_d(s + t) - f_d(s)}{t} \,, \qquad
    \partial_{-} f_d(s) = \lim_{t \uparrow 0^-} \frac{f_d(s + t) - f_d(s)}{t} \,,
\end{equation}
where the above limits exist and are finite by the convexity of $f_d(s)$ and monotone-convergence theorem. The subdifferential $\partial f_d(s)$ at a point $s$ is then given by the interval $\partial f_d(s) =  [\partial_{-} f_d(s),\partial_{+} f_d(s)]$ \citep[Theorem~23.4]{rockafellar1970convex}.

Using the same notation for $f$ (instead of $f_d$), by the definition of $\partial_{+} f(s), \partial_{-} f(s)$, for any $\varepsilon > 0$ small enough, there exist $s^+_\varepsilon, s^-_\varepsilon \in O$ such that:
\begin{equation}\label{eq:pineq}
 \frac{f(s^+_\varepsilon)-f(s)}{s^+_\varepsilon-s} \leq \partial_{+} f(s) + \frac{\varepsilon}{2},
\end{equation}
and
\begin{equation}\label{eq:minineq}
    \frac{f(s^-_\varepsilon)-f(s)}{s^-_\varepsilon-s} \geq \partial_{-} f(s) - \frac{\varepsilon}{2} \,.
\end{equation}

By the convexity of $f_d(s)$, we further have
\begin{equation}\label{eq:convp}
    \partial_{+} f_d(s) \leq  \frac{f_d(s^+_\epsilon)-f_d(s)}{s^+_\varepsilon-s} \,,
\end{equation}
and
\begin{equation}\label{eq:convm}
    \partial_{-} f_d(s) \geq  \frac{f_d(s^-_\epsilon)-f_d(s)}{s^-_\varepsilon-s} \,.
\end{equation}

Define the following sequence of events
\begin{align}
    \mathcal{E}^+_{d,\varepsilon} &: \abs{\frac{f_d(s^+_\varepsilon)-f(s^+_\varepsilon)}{s^+_\varepsilon-s}} \leq \frac{\varepsilon}{4} \,, \\
    \mathcal{E}^-_{d,\varepsilon} &: \abs{\frac{f_d(s^-_\varepsilon)-f(s^-_\varepsilon)}{s^-_\varepsilon-s}} \leq \frac{\varepsilon}{4} \,, \\
    \mathcal{E}^s_{d,\varepsilon} &: \abs{f_d(s)-f(s)} \leq \frac{\varepsilon}{4} \operatorname{min}\left({\abs{s^-_\varepsilon-s}, \abs{s^+_\varepsilon-s}}\right) \,,
\end{align}

By the point-wise a.s convergence of $f_d(s)$ to $f(s)$, we have
\begin{equation}
    \lim_{d_0\rightarrow \infty} \Pr[\cap_{d\geq d_0} \mathcal{E}^+_{d,\varepsilon}] = 1 \,, \quad \lim_{d_0\rightarrow \infty} \Pr[\cap_{d\geq d_0} \mathcal{E}^-_{d,\varepsilon}]  = 1 \,, \quad \lim_{d_0\rightarrow \infty} \Pr[\cap_{d\geq d_0} \mathcal{E}^s_{d,\varepsilon}]  = 1 \,. 
\end{equation}

Applying a union-bound to the complements $[\cap_{d\geq d_0} \mathcal{E}^+_{d,\varepsilon}]^\complement, [\cap_{d\geq d_0} \mathcal{E}^-_{d,\varepsilon}]^\complement, [\cap_{d\geq d_0} \mathcal{E}^s_{d,\varepsilon}]^\complement$, we obtain that for any $\varepsilon > 0$ 
\begin{equation}\label{eq:asinc}
    \Pr[\cap_{d\geq d_0} [\mathcal{E}^+_{d,\varepsilon} \cap \mathcal{E}^-_{d,\varepsilon} \cap \mathcal{E}^s_{d,\varepsilon}]] \rightarrow 1 \,,
\end{equation}
as $d_0 \rightarrow \infty$. 

Next, suppose that the events $\mathcal{E}^+_{d,\varepsilon}, \mathcal{E}^-_{d,\varepsilon}, \mathcal{E}^s_{d,\varepsilon}$ hold for some $d \in \mathbb{N}$. Then, adding and subtracting $\frac{f_d(s^+_\varepsilon)-f(s^+_\varepsilon)}{s^+_\varepsilon-s}$ and $\frac{f_d(s)-f(s)}{s^+_\varepsilon-s}$ in the RHS of \cref{eq:convp} and applying triangle inequality yields:
\begin{equation}
    \partial_{+} f_d(s) \leq  \frac{f(s^+_\epsilon)-f(s)}{s^+_\varepsilon-s}+\abs{\frac{f_d(s^+_\epsilon)-f(s_\epsilon^+)}{s^+_\varepsilon-s}}+\abs{\frac{f_d(s)-f(s)}{s^+_\varepsilon-s}} \leq \frac{f(s^+_\epsilon)-f(s)}{s^+_\varepsilon-s}+\frac{\varepsilon}{2} \,.
\end{equation}
Combining the above with \cref{eq:pineq}, we obtain:
\begin{equation}
    \partial_{+} f_d(s) \leq \partial_{+} f(s)+\varepsilon.
\end{equation}

Analogously, \cref{eq:convm,eq:minineq} imply that:
\begin{equation}
    \partial_{-} f_d(s) \leq \partial_{-} f(s)+\varepsilon \,.
\end{equation}

Recalling the relations, $\partial f_d(s) =  [\partial_{-} f_d(s),\partial_{+} f_d(s)]$ and $\partial f(s) =  [\partial_{-} f(s),\partial_{+} f(s)]$ we obtain:
\begin{equation}
    \partial f_d(s) \in \partial f(s)+\varepsilon B \,.
\end{equation}

Since the above event holds under 
$\mathcal{E}^+_{d,\varepsilon} \cap \mathcal{E}^-_{d,\varepsilon} \cap \mathcal{E}^s_{d,\varepsilon}$, \cref{eq:asinc} implies~\cref{eq:grad_inc}.
\end{proof}

We are ready to conclude the proof of our main result. 

\begin{proof}[\Cref{thm:main}]
We prove each part separately.

\textit{Part 1.} 
We establish almost sure convergence through a sandwich argument.
For any $\varepsilon > 0$, we show that the sequence $\minriskfun_d$ is eventually bounded between $\minriskfun \pm \varepsilon$ almost surely.
By~\Cref{thm:lower_b}, with probability $1$,
there exists $d_\varepsilon \in \mathbb{N}$ satisfying
\begin{equation}\label{eq:lower_bound}
\minriskfun_d(\omega) \geq \minriskfun - \varepsilon \quad \forall d > d_\varepsilon \,.
\end{equation}
Similarly, by~\Cref{prop:physicality-fixed-point}, with probability $1$, there exists $\tilde{d}_\varepsilon \in \mathbb{N}$ satisfying
\begin{equation}\label{eq:upper_bound}
    \minriskfun_d(\omega) \leq \minriskfun + \varepsilon \quad \forall d > \tilde{d}_\varepsilon(\omega) \,.
\end{equation}
Then, taking $D_\varepsilon= \max\{d_\varepsilon, \tilde{d}_\varepsilon\}$, we have:
\begin{equation}
    |\minriskfun_d - \minriskfun| \leq \varepsilon \quad \forall d > D_\varepsilon \,.
\end{equation}

This implies that for all $\varepsilon > 0$
\begin{equation}
    \Pr[
        \limsup_{d \to \infty} |\minriskfun_d - \minriskfun| \geq \varepsilon
    ] = 0 \,,
\end{equation}
which establishes the almost sure convergence in~\cref{eq:min-risk-limit-def}.

\textit{Part 2.} 
For any finite dimension $d$, define the perturbed minimum for $s\in\RR$
\begin{equation}
    \Psi_d(s) = \min_{\w \in \mathcal{K}_{a,b}} \Big\{
        \riskfun_d(\w) + s \psi\Big(\frac{1}{d} \langle\w, \wteacher \rangle, \frac{1}{d} \norm{\w}_2^2\Big)
    \Big\} \,.
\end{equation}
By Danskin's theorem \citep{Bertsekas01031997}, for $s = 0$, the subdifferential $\partial \Psi_d(0)$ contains the set
\begin{equation}
    \qty{
        \psi\Big(\frac{1}{d} \langle\what, \wteacher \rangle, \frac{1}{d} \norm{\what}_2^2\Big) : \what \in \argmin_{\w \in \mathcal{K}_{a,b}} \riskfun_d(\w)
    } \,.
\end{equation}

The limiting function $\Psi(s)$ defined in~\cref{eq:def_f} is concave as an infimum of affine functions in $s$ again because of Danskin's theorem. By assumption, it is differentiable at $s = 0$.

By~\Cref{thm:conv-derivative}, for any $\varepsilon > 0$, with probability $1$, there exists $d_\psi$ satisfying
\begin{equation}
    \psi\Big(\frac{1}{d} \langle\wminimizer, \wteacher \rangle, \frac{1}{d} \norm{\wminimizer}_2^2\Big) \in \Big[\eval{\dv{s} \Psi(s)}_{s=0} - \varepsilon, \eval{\dv{s} \Psi(s)}_{s=0} + \varepsilon\Big]
\end{equation}
for all $d > d_\psi$. 
The fact that $\varepsilon$ is arbitrary establishes~\cref{eq:def_psi}.
\end{proof}

\newpage
\appendix

\crefalias{section}{appendix}
\newpage

\section{Connection between Weak Convexity and Lipshitzness of Proximals}\label{sec:app:from-weak-convex-to-lipshitz}

This appendix shows how~\Cref{ass:weak_convexity} is both necessary and sufficient for the existence of Lipschitz-continuous proximal selections required in our framework, by building up the tools and preliminary results to prove~\Cref{thm:main_proximal_characterization,cor:scaled_proximal}.

As this appendix may be of broader interest beyond the setting of the main paper, we slightly modify the notation. We will work, whenever possible, with a generic function $f \, :\, \RR \rightarrow \RR$ and denote $E_f \subseteq \mathbb{R}$ as the set of inputs $\delta\in \RR$ such that $\mathcal{P}_f(\cdot)$ is non-empty, where $\mathcal{P}_f(\cdot)$ is here defined as :

\begin{equation}
    \mathcal{P}_f(\delta) \defeq \argmin_{r\in  \RR} \frac{(r-\delta)^2}{2}+f(r).
\end{equation}

Similarly, the Moreau enveloppe is here defined as :

\begin{equation}
    \mathcal{M}_f(\delta) \defeq \min_{r\in  \RR} \frac{(r-\delta)^2}{2}+f(r).
\end{equation}

Throughout, we denote by $r_{\min}: \mathbb{R} \rightarrow \mathbb{R}$ an arbitrary proximal selection defined on $E_f$, i.e. $\forall \, \delta \in E_f, r_{\min}(\delta) \in \mathcal{P}_f(\delta)$. When the proximal operator is singleton-valued, the proximal selection is uniquely determined.

Before starting we also recall a fundamental property of convex functions that will be used repeatedly.

\begin{lemma}[Properties of Convex Functions]\label{lem:convex-properties}
Let $g: \RR \to \RR \cup \{\infty\}$ be a convex function. The set of points where $g$ is finite is an interval $J$. On the interior $\mathring{J}$, the left and right derivatives $g'_-$ and $g'_+$ exist everywhere and satisfy:
\begin{enumerate}[label=(\roman*)]
    \item $g'_-$ and $g'_+$ are non-decreasing functions with $g'_- \leq g'_+$.
    \item $g'$ is defined almost everywhere on $J$ and is non-decreasing.
    \item For all $x \in \mathring{J}$, $g'_-(x) = g'(x^-)$ and $g'_+(x) = g'(x^+)$.
    \item $g$ is differentiable at $\delta \in \mathring{J}$ if and only if $g'_-(\delta) = g'_+(\delta)$, which holds if and only if $g'$ is continuous at $\delta$.
\end{enumerate}
\end{lemma}

\subsection{Existence and Basic Properties of Proximal Operators}

Although it is a necessary condition, asking that $f$ is lower semi continuous (l.s.c.) does not guarantee non-emptiness of the proximal operator of $f$. Indeed, $r\mapsto \exp(-r^2)-\frac{r^2}{2}$ is l.s.c. but its proximal operator at $\delta=0$ is empty because the minimizer has fled to infinity. The purpose of the prox-coercive property that we impose on $f$ is to prevent this from happening.

We first recall key properties of proximal operators for weakly convex functions.

\begin{proposition}\label{prop:existence_proximal}
Suppose $f$ is lower semi-continuous and prox-coercive. For $V \in [0,1]$, $E_{Vf} = \mathbb{R}$ and in particular the Moreau envelope of $Vf$ is always finite. Furthermore, for any $\delta \in \mathbb{R}$, $\mathcal{P}_{Vf}(\delta)$ is a compact set.
\end{proposition}

\begin{proof}[\Cref{prop:existence_proximal}]
Fix $\delta \in \mathbb{R}$ and $V \in (0,1]$. Consider a minimizing sequence $(r_n)_{n \in \mathbb{N}} \subset \mathbb{R}$ such that
$$\frac{(\delta - r_n)^2}{2} + Vf(r_n) \to \mathcal{M}_{Vf}(\delta) \in [-\infty, \infty).$$

Suppose by contradiction that $(|r_n|)_n$ is unbounded. Then it has a subsequence with $|r_{n_k}| \to \infty$. Since $f$ is prox-coercive,
$\limsup_{k \to \infty} \left[(\delta - r_{n_k})^2/ 2 + f(r_{n_k})\right] = \infty$.
However, for any $n \in \mathbb{N}$,
$$\frac{(\delta - r_n)^2}{2} + Vf(r_n) = V\left[\frac{(\delta - r_n)^2}{2} + f(r_n)\right] + (1-V)\frac{(\delta - r_n)^2}{2}.$$

Since $V \in (0,1]$ and the second term is non-negative, taking $\limsup$ along the subsequence yields a contradiction to the convergence of the left-hand side.

Thus $(r_n)_n$ is bounded. Let $r$ be an accumulation point. By lower semi-continuity, taking $\liminf$ along a convergent subsequence yields
$(\delta - r)^2 / 2 + Vf(r) \leq \mathcal{M}_{Vf}(\delta)$.
The reverse inequality holds by definition of the Moreau envelope, establishing that $r \in \mathcal{P}_{Vf}(\delta)$. 

The same argument shows $\mathcal{P}_{Vf}(\delta)$ is closed and bounded, hence compact.
\end{proof}

\begin{proposition}\label{prop:convexity_moreau}
The following function is convex:
$$g : \delta \in \mathbb{R} \mapsto \frac{\delta^2}{2} - \mathcal{M}_f(\delta) \in \mathbb{R} \cup \{\infty\}.$$

Hence, $\mathcal{M}_f$ is almost everywhere twice differentiable on the interval $I_f := \{\delta \in \mathbb{R} \mid \mathcal{M}_f(\delta) > -\infty\}$.

In addition, for $\delta \in I_f$,
$$\mathcal{P}_f(\delta) \subset [g'_-(\delta), g'_+(\delta)].$$

If $\mathcal{M}_f$ is differentiable at $\delta \in E_f$ (i.e., $\mathcal{M}'_f$ is continuous at $\delta$), then $\mathcal{P}_f(\delta)$ is a singleton.

Furthermore, any proximal selection of $f$ is non-decreasing on $E_f$. If there exists a non-trivial interval $I \subset I_f$ such that $I \subset E_f$, this selection is continuous at $\delta$ if and only if $\mathcal{M}_f$ is differentiable at $\delta$, and differentiable almost everywhere on $I$.
\end{proposition}

\begin{proof}[\Cref{prop:convexity_moreau}]
We can rewrite $g(\delta)$ as
$$g(\delta) = \sup_{r \in \mathbb{R}} \left[\delta r - \frac{r^2}{2} - f(r)\right].$$

Since $\delta \mapsto \delta r - \frac{r^2}{2} - f(r)$ is convex for any $r \in \mathbb{R}$, $g$ is convex as a supremum of convex functions. By~\Cref{lem:convex-properties}, $g$ is differentiable almost everywhere on $I_f$, hence so is $\mathcal{M}_f$.

Now fix $\delta \in I_f$ and suppose $\mathcal{P}_f(\delta)$ is non-empty. Take $r_{\min}(\delta) \in \mathcal{P}_f(\delta)$. For $r \in \mathbb{R}$ and $\delta' \neq \delta$,
$$\frac{\delta'^2}{2} - \mathcal{M}_f(\delta') \geq \frac{\delta'^2}{2} - \frac{(\delta' - r)^2}{2} - f(r) = \frac{\delta^2}{2} - \frac{(\delta - r)^2}{2} - f(r) + (\delta' - \delta)r.$$

At $r = r_{\min}(\delta)$, this gives
$$g(\delta') - g(\delta) \geq (\delta' - \delta)r_{\min}(\delta).$$

Dividing by $\delta' - \delta \neq 0$ yields $r_{\min}(\delta) \in [g'_-(\delta), g'_+(\delta)]$.

By~\Cref{lem:convex-properties}, $g$ is differentiable at $\delta$ if and only if $g'_-(\delta) = g'_+(\delta)$, which implies $\mathcal{P}_f(\delta)$ is a singleton.

For any proximal selection $r_{\min}$ defined on $E_f$ and $\delta < \delta'$ in $E_f$,
$$r_{\min}(\delta) \leq g'_+(\delta) = g'(\delta^+) \leq g'(\delta'^-) = g'_-(\delta') \leq r_{\min}(\delta'),$$
where we used~\Cref{lem:convex-properties}. Thus $r_{\min}$ is non-decreasing.

For continuity, suppose $\delta \in \mathring{I}$ and $\mathcal{M}_f$ is differentiable at $\delta$. Since $\mathcal{M}'_f$ is continuous at $\delta$, there exist sequences $(\delta_{n,\pm})_n \subset I$ with $\delta_{n,\pm} \to \delta^\pm$ and $\mathcal{M}_f$ differentiable at each $\delta_{n,\pm}$. Then
$$r_{\min}(\delta_{n,\pm}) = \delta_{n,\pm} - \mathcal{M}'_f(\delta_{n,\pm}) \to \delta - \mathcal{M}'_f(\delta) = r_{\min}(\delta),$$
establishing continuity. Conversely, the above chain of inequalities yield at $\delta\in \mathring{I}$

$$r_{\min}(\delta^-) \leq g'_+(\delta^-) \leq g'_-(\delta^+) \leq r_{\min}(\delta^+).$$

If $r_{\min}$ is continuous at $\delta$ then so are $g'$ and $\mathcal{M}_f'$. In particular, $\mathcal{M}_f$ is differentiable at $\delta$.

Almost everywhere differentiability follows from Lebesgue's theorem on monotone functions.
\end{proof}

\begin{remark}
In particular, if $E_f = \mathbb{R}$, such as when $f$ is lower semi-continuous and prox-coercive (by Proposition~\ref{prop:existence_proximal}), then any proximal selection of $f$ is non-decreasing.
\end{remark}

\subsection{Characterization via Chord Slope Functions}

\begin{proposition}\label{prop:si_characterization}
Take $r_{\min}$ to be a proximal selection of $f$ defined on $E_f$. Then we have
\begin{equation}
    E_f = \{\delta \in \mathbb{R} \mid \exists r_0 \in \mathbb{R}, \, s_f(r_0) \leq \delta \leq i_f(r_0)\}, \label{eq:app:si:Ef}
\end{equation}
and for any $r_0, \delta \in \mathbb{R}$,
\begin{equation}
    r_0 \in \mathcal{P}_f(\delta) \iff s_f(r_0) \leq \delta \leq i_f(r_0), \label{eq:app:si:r0}
\end{equation}
where $s_f$ and $i_f$ come from~\Cref{def:app:si}.
\end{proposition}

\begin{proof}[\Cref{prop:si_characterization}]
For any $r_0, \delta \in \mathbb{R}$, by the definition of $\mathcal{P}_f(\delta)$,
$$r_0 \in \mathcal{P}_f(\delta) \iff \forall r \in \mathbb{R}, \quad \delta(r - r_0) \leq f(r) - f(r_0) + \frac{r + r_0}{2}(r - r_0).$$

Separating the cases $r > r_0$ and $r < r_0$, this is equivalent to
$$\forall r > r_0, \quad \delta \leq \frac{f(r) - f(r_0)}{r - r_0} + \frac{r + r_0}{2} \quad \text{and} \quad \forall r < r_0, \quad \delta \geq \frac{f(r) - f(r_0)}{r - r_0} + \frac{r + r_0}{2}.$$

By the definitions of supremum and infimum, we obtain~\cref{eq:app:si:r0}.
To establish~\cref{eq:app:si:Ef} we notice the following. We have that $\delta \in E_f$ if and only if $\mathcal{P}_f(\delta) \neq \emptyset$, which occurs if and only if there exists $r_0 \in \mathbb{R}$ with $s_f(r_0) \leq \delta \leq i_f(r_0)$.
\end{proof}

\begin{remark}
\Cref{prop:si_characterization} implies that for $\delta \in E_f$, the left chords of $f$ at $r_0$ are uniformly bounded above, and the right chords are uniformly bounded below. It also implies
$$f'_-(r_0) := \limsup_{r \to r_0^-} \frac{f(r) - f(r_0)}{r - r_0} \leq f'_+(r_0) := \liminf_{r \to r_0^+} \frac{f(r) - f(r_0)}{r - r_0}.$$
\end{remark}

\begin{proposition}\label{prop:weak_convex_si}
The following equivalence holds:
$$\forall r\in \mathbb{R}, \quad s_f(r) \leq i_f(r) \iff f \text{ is } 1\text{-weakly convex},$$
where $s_f$ and $i_f$ come again from~\Cref{def:app:si}.
Furthermore, when this is the case, $s_f = f'_- + \mathrm{id}$ and $i_f = f'_+ + \mathrm{id}$ are non-decreasing.
\end{proposition}

\begin{proof}[\Cref{prop:weak_convex_si}]
Define $F : x \mapsto f(x) + \frac{x^2}{2}$. For $r \neq r_0$,
$$\frac{f(r) - f(r_0)}{r - r_0} + \frac{r + r_0}{2} = \frac{F(r) - F(r_0)}{r - r_0}.$$

If $F$ is convex, then by~\Cref{lem:convex-properties},
$s_f(r_0) = F'_-(r_0) = F'(r_0^-) \leq F'(r_0^+) = F'_+(r_0) = i_f(r_0)$.
Conversely, if $s_f(r_0) \leq i_f(r_0)$ for all $r_0 \in \mathbb{R}$, then for $r < r_0 < r'$,
$$\frac{F(r) - F(r_0)}{r - r_0} \leq s_f(r_0) \leq i_f(r_0) \leq \frac{F(r') - F(r_0)}{r' - r_0}.$$

This chain of inequalities characterizes convexity of $F$.

When $F$ is convex, we observe that $s_f = F'_- = f'_- + \mathrm{id}$ and $i_f = F'_+ = f'_+ + \mathrm{id}$ are indeed non-decreasing.
\end{proof}

\begin{corollary}\label{cor:strict_weak_convex}
Suppose $f$ is $1$-weakly convex. The following equivalences hold:
\begin{equation}
    s_f \text{ is increasing} \iff f \text{ is strictly } 1\text{-weakly convex} \iff i_f \text{ is increasing},
\end{equation}
where $s_f$ and $i_f$ are from~\Cref{def:app:si}.
Furthermore, $s_f$ is constant on a non-trivial interval if and only if $i_f$ is also constant, with the same value, on that interval.
\end{corollary}

\begin{proof}[\Cref{cor:strict_weak_convex}]
By~\Cref{prop:weak_convex_si}, $s_f = F'_-$ and $i_f = F'_+$ where $F : r \mapsto f(r) + \frac{r^2}{2}$ is convex.

Suppose $F$ is strictly convex. Fix $r_1 < r_2 < r_3 < r_4$ in $\mathbb{R}$. Strict convexity yields
$$\frac{F(r_2) - F(r_1)}{r_2 - r_1} < \frac{F(r_3) - F(r_2)}{r_3 - r_2} < \frac{F(r_4) - F(r_2)}{r_4 - r_2} < \frac{F(r_4) - F(r_3)}{r_4 - r_3}.$$

Taking the limit $r_1 \to r_2^-$ yields $s_f(r_2) \leq \frac{F(r_3) - F(r_2)}{r_3 - r_2} < \frac{F(r_4) - F(r_2)}{r_4 - r_2}$. Then taking $r_3 \to r_4^-$ gives $\frac{F(r_4) - F(r_2)}{r_4 - r_2} \leq s_f(r_4)$. Combined, we deduce $s_f(r_2) < s_f(r_4)$, so $s_f$ is increasing. The proof for $i_f$ is analogous.

Conversely, if $F$ is not strictly convex, there exists a non-trivial open interval $J$ on which $F$ is linear. On $J$, $F'$ is constant, hence $s_f|_J$ and $i_f|_J$ are constant and equal.
\end{proof}

\begin{remark}
If $f$ is strictly $1$-weakly convex, then $r \mapsto f(r) + \frac{r^2}{2}$ has at most one minimum, and hence $\mathcal{P}_f$ is singleton-valued or empty. If $E_f = \mathbb{R}$, then $\mathcal{P}_f$ is singleton-valued everywhere.
\end{remark}

\subsection{Coercivity and Continuity of Proximal Selections}

\begin{proposition}\label{prop:locally_lower_bounded}
Suppose there exists $\delta \in \mathbb{R}$ such that $\mathcal{M}_f(\delta) > -\infty$. Then $f$ is locally lower-bounded.
\end{proposition}

\begin{proof}[\Cref{prop:locally_lower_bounded}]
Suppose by contradiction that $f$ is not locally lower-bounded at some $r_\infty \in \mathbb{R}$. 
Thus, for any $n \in \mathbb{N}^*$, there exists $r_n \in (r_\infty - \frac{1}{n}, r_\infty + \frac{1}{n})$ with $f(r_n) \leq -n$. Thus $r_n \to r_\infty$ and
$$\mathcal{M}_f(\delta) \leq \frac{(r_n - \delta)^2}{2} + f(r_n) \leq \frac{(r_n - \delta)^2}{2} - n \to -\infty,$$
which contradicts $\mathcal{M}_f(\delta) > -\infty$.
\end{proof}

\begin{remark}
In particular, if there exists $\delta \in \mathbb{R}$ such that $\mathcal{P}_f(\delta) \neq \emptyset$, then $f$ is locally lower-bounded.
\end{remark}

\begin{proposition}\label{prop:coercive_selection}
Suppose there exists a sequence $(\delta_n)_{n \in \mathbb{N}} \subset \mathbb{R}$ such that for all $n \in \mathbb{N}$, $\mathcal{P}_f(\delta_n) \neq \emptyset$ and $|\delta_n| \to \infty$.

Then any proximal selection $r_{\min}$ of $f$ satisfies $|r_{\min}(\delta_n)| \to \infty$.
\end{proposition}

\begin{proof}[\Cref{prop:coercive_selection}]
Suppose by contradiction that $|r_{\min}(\delta_n)|$ does not diverge. Then there exists $M > 0$ such that a subsequence (still denoted $(\delta_n)$) satisfies $|r_{\min}(\delta_n)| \leq M$ for all $n$ and converges to some $r_\infty \in \mathbb{R}$.

By~\Cref{prop:locally_lower_bounded}, $f$ is locally lower-bounded. Thus there exists an open set $U \subset \mathbb{R}$ containing $r_\infty$ with $\inf_U f > -\infty$. Without loss of generality, assume $r_{\min}(\delta_n) \in U$ for all $n$.

For $n$ such that $|\delta_n| \geq M$, by the triangle inequality,
$$\mathcal{M}_f(\delta_n) \geq \frac{(|\delta_n| - M)^2}{2} + \inf_U f.$$

However, for any $r \in \mathbb{R}$,
$$\mathcal{M}_f(\delta_n) \leq \frac{(\delta_n - r)^2}{2} + f(r).$$

Setting $r_n = \mathrm{sign}(\delta_n)(M+1)$ and defining $M' = \frac{(M+1)^2}{2} + \max(f(M+1), f(-(M+1)))$, we obtain
$$M' \geq \mathcal{M}_f(\delta_n) - \frac{\delta_n^2}{2} + r_n \delta_n \geq \frac{M^2}{2} - M|\delta_n| + r_n \delta_n + \inf_U f = |\delta_n| + \frac{M^2}{2} + \inf_U f \to \infty,$$
which is a contradiction.
\end{proof}

\begin{corollary}\label{cor:continuity_equiv}
Suppose $E_f = \mathbb{R}$. Let $r_{\min}$ be a proximal selection of $f$. Then
$$r_{\min} \in C^0(\mathbb{R}, \mathbb{R}) \iff f \text{ is strictly } 1\text{-weakly convex}.$$

Furthermore, when this is the case, $\mathcal{P}_f$ is singleton-valued and $r_{\min}$ is an onto function.
\end{corollary}

\begin{proof}[\Cref{cor:continuity_equiv}]
With $E_f = \mathbb{R}$, $\mathcal{M}_f$ is finite everywhere and Proposition~\ref{prop:convexity_moreau} shows $r_{\min}$ is non-decreasing, and Proposition~\ref{prop:coercive_selection} shows $|r_{\min}|$ is coercive. Hence $r_{\min}$ is continuous if and only if it is onto.

Suppose $r_{\min}$ is onto. Proposition~\ref{prop:si_characterization} guarantees $s_f \leq i_f$. By Proposition~\ref{prop:weak_convex_si}, this implies $f$ is $1$-weakly convex. Since $r_{\min}$ is continuous, Proposition~\ref{prop:convexity_moreau} shows $\mathcal{M}_f$ is differentiable and thus $\mathcal{P}_f$ is singleton-valued.

Suppose by contradiction that $f$ is not strictly $1$-weakly convex. By Corollary~\ref{cor:strict_weak_convex}, there exist $r_1 < r_2$ with $s_f(r_1) = s_f(r_2) = s_f(r) = i_f(r)$ for $r \in [r_1, r_2]$. In particular, $\delta = s_f(r_1)$ satisfies $s_f(r_1) \leq \delta \leq i_f(r_1)$ and $s_f(r_2) \leq \delta \leq i_f(r_2)$. By Proposition~\ref{prop:si_characterization}, $r_1, r_2 \in \mathcal{P}_f(\delta)$, which contradicts singleton-valuedness.

Conversely, suppose $f$ is strictly $1$-weakly convex. By Remark following Corollary~\ref{cor:strict_weak_convex}, $\mathcal{P}_f$ is singleton-valued. Define
$$\delta : r_0 \mapsto \frac{s_f(r_0) + i_f(r_0)}{2}.$$

By Proposition~\ref{prop:weak_convex_si}, $s_f(r_0) \leq \delta(r_0) \leq i_f(r_0)$. By Proposition~\ref{prop:si_characterization}, $r_0 \in \mathcal{P}_f(\delta(r_0)) = \{r_{\min}(\delta(r_0))\}$. Thus $r_0 \in \mathrm{Im}(r_{\min})$, proving $r_{\min}$ is onto.
\end{proof}

\begin{remark}
In particular, when $f$ is lower semi-continuous and prox-coercive, this equivalence holds, since these hypotheses imply $E_f = \mathbb{R}$ by~\Cref{prop:existence_proximal}.
\end{remark}

\begin{remark}
If $f$ is strictly $1$-weakly convex, $r_{\min}$ is the only function such that
$\forall \delta \in \mathbb{R}, s_f(r_{\min}(\delta)) \leq \delta \leq i_f(r_{\min}(\delta))$.
\end{remark}

\subsection{Lipschitz Continuity from Weak Convexity}

\begin{lemma}\label{lem:lipschitz_from_weak_convex}
Suppose $f$ is $V_0$-weakly convex for some $V_0 > 1$. Then, $E_f=\RR$. Furthermore, let $r_{\min}$ be a proximal selection of $f$. Then $r_{\min}$ is $\frac{V_0}{V_0 - 1}$-Lipschitz continuous.
\end{lemma}

\begin{proof}[\Cref{lem:lipschitz_from_weak_convex}]
Define $G : r \mapsto V_0 f(r) + \frac{r^2}{2}$, which is convex by $V_0$-weak convexity. For any $\delta \in \mathbb{R}$, the function $F_\delta : r \mapsto f(r) + \frac{(\delta - r)^2}{2}$ can be written as
\begin{equation}
    F_\delta(r) = F_0(r) + \left(-\delta r + \frac{\delta^2}{2}\right) \,, \qquad \text{with} \qquad 
    F_0(r) = \frac{1}{V_0}G(r) + \left(1 - \frac{1}{V_0}\right)\frac{r^2}{2} \,.
\end{equation}
Since $V_0 > 1$, both terms on the right are convex, hence $F_0$ is convex, and consequently $F_\delta$ is convex.
Furthermore, if $r_0\in \RR$ is such that $G$ is differentiable at $r_0$, we have that, for any $r\in \RR, \: G(r) \geq G(r_0) + G'(r_0)(r-r_0)$. Thus, $F_\delta$ is coercive in $r$, and $f$ is prox-coercive. Since it is weakly convex, it also is continuous and hence $E_f=\RR$.

Since $r_{\min}(\delta)$ minimizes the convex function $F_\delta$, the first-order optimality condition gives $0 \in \partial F_\delta(r_{\min}(\delta))$. From the expressions above,
$$\partial F_\delta(r) = \frac{1}{V_0}\partial G(r) + \left(1 - \frac{1}{V_0}\right)r - \delta.$$

Therefore, at $r = r_{\min}(\delta)$,
$$V_0 \delta - (V_0 - 1)r_{\min}(\delta) \in \partial G(r_{\min}(\delta)) = [G'_-(r_{\min}(\delta)), G'_+(r_{\min}(\delta))].$$

For $\delta < \delta'$, using that $r_{\min}$ is non-decreasing (\Cref{prop:convexity_moreau}) and $G$ is convex (hence $\partial G$ is monotone),
$$V_0 \delta - (V_0 - 1)r_{\min}(\delta) \leq G'_+(r_{\min}(\delta)) \leq G'_-(r_{\min}(\delta')) \leq V_0 \delta' - (V_0 - 1)r_{\min}(\delta').$$

Rearranging yields
$$0 \leq r_{\min}(\delta') - r_{\min}(\delta) \leq \frac{V_0}{V_0 - 1}(\delta' - \delta).$$
which is the definition of Lipschitzness that we were looking for.
\end{proof}

Before proving the reciprocal, we recall a well known characterization of convex functions over intervals in our specific case.

\begin{lemma}\label{lem:convex-lip-continuous}
    Let $F \, :\, \RR \rightarrow \RR$ be any function. 

    \begin{equation}
        F \text{ is convex} \iff F \text{ is locally Lipschitz continuous with } F' \text{ non-decreasing}
    \end{equation}
\end{lemma}

\begin{remark}
    A Lipschitz continuous function on a compact set is differentiable almost everywhere on this compact set. Thus, in the above equivalence, $F'$ non-decreasing must be understood as $F'$ is non decreasing over the set where it is defined. 
\end{remark}

\begin{proof}[\Cref{lem:convex-lip-continuous}]
    Take $a<b\in \RR$.
    Suppose $F$ is convex. We know that $F$ is almost everywhere differentiable with $F'$ non-decreasing.
    For $r'<r$ elements of $[a,b]$,
    \begin{equation}
        F'_-(a)\leq\frac{F(r)-F(r')}{r-r'}\leq F'_+(b).
    \end{equation}

    Thus, $F$ is $\max (|F'_+(a)|,|F'_-(b)|)$-Lipschitz continuous on $[a,b]$.
    Conversely, if $F$ is Lipschitz continuous on $[a,b]$, then it is absolutely continuous on $[a,b]$ and for $r'<r\in[a,b]$,
    \begin{equation}
        F(r)-F(r') =\int_{r'}^rF'.
    \end{equation}
    
    Now, since $F'$ is non-decreasing, then
    \begin{equation}\label{eq:sufficient-ineq-slope-derivative-convex}
        F'(r'^+)=\inf_{[r',r]} F'\leq \frac{F(r)-F(r')}{r-r'}\leq  \sup_{[r',r]}F' = F'(r^-) \, .
    \end{equation}

    This suffices to show that $F$ is convex. Indeed, take $r''\in [a,b]$ such that $r'' <r'$. Applying \cref{eq:sufficient-ineq-slope-derivative-convex} to both pairs $(r'',r')$ and $(r',r)$, and using the fact that $F'(\cdot^-)\leq F'(\cdot^+)$ because $F'$ is non-decreasing, yields
    \begin{equation}
        \frac{F(r')-F(r'')}{r'-r''} \leq F'(r'^-) \leq F'(r'^+) \leq \frac{F(r)-F(r')}{r-r'}.
    \end{equation}
    This is a well known characterization of convex functions.
\end{proof}

\begin{corollary}\label{cor:lipschitz_equiv}
    Suppose that $E_f=\RR$. Let $r_{\min}$ be a proximal selection of $f$. For $L\in (1,\infty)$, we have
    \begin{equation}
        r_{\min} \text{ is $L$-Lipschitz continuous} \iff f \text{ is $\frac{L}{L-1}$-weakly convex},
    \end{equation}
    and $r_{\min}$ is $1$-Lipschitz continuous if and only if $f$ is convex.
\end{corollary}

\begin{proof}[\Cref{cor:lipschitz_equiv}]
    Suppose $r_{\min}$ is $L$-Lipschitz continuous.
    In particular, it is continuous and the ~\Cref{cor:continuity_equiv} yields the fact that $f$ is strictly $1$-weakly convex. 
    
    From \Cref{lem:convex-lip-continuous} we know that $r\mapsto f(r)+\frac{r^2}{2}$ and $r\mapsto-\frac{r^2}{2}$ are locally Lipschitz continuous, therefore $f$ has the same property. 
    
    Let us denote by $F : r \mapsto \frac{L}{L-1}f(r)+\frac{r^2}{2}$. Our aim is to show that $F$ is convex, \textit{i.e.} locally Lipschitz continuous with non-decreasing derivative by \Cref{lem:convex-lip-continuous}. We already know that $F$ is locally Lipschitz continuous, since it is the case of $f$.

    By ~\Cref{cor:continuity_equiv,prop:convexity_moreau} and continuity, $r_{\min}$ is non-decreasing and onto. Thus it suffices to check that $G\defeq F'\circ r_{\min}$ is non-decreasing wherever it is defined.

    Take $\delta\in \RR$ such that $G$ is defined at $\delta$. Then, $f$ is differentiable at $r_{\min}(\delta)$, which minimizes the function $r \in \RR \mapsto \frac{(\delta-r)^2}{2}+f(r)$.

    Therefore, the extremality condition yields 
    \begin{equation}
        \delta-r_{\min}(\delta)=\frac{\partial f}{\partial r}(r_{\min}(\delta)),
    \end{equation}
    and thus 
    \begin{equation}\label{eq:G-for-monotone-derivative}
        G(\delta)=\frac{L\delta}{L-1}-\frac{r_{\min}(\delta)}{L-1}.
    \end{equation}

    However, for $\delta<\delta'$ real numbers,
    \begin{equation}
        0\leq r_{\min}(\delta')-r_{\min}(\delta)\leq L(\delta'-\delta),
    \end{equation}
    and thus 
    \begin{equation}\label{eq:lip-eq-prox}
        -r_{\min}(\delta)\leq -r_{\min}(\delta')+L(\delta'-\delta)
    \end{equation}

    Fix $\delta_0<\delta_1$ in the domain of $G$.

    Using~\cref{eq:G-for-monotone-derivative,eq:lip-eq-prox},
    \begin{equation}
        G(\delta)\leq \frac{L\delta}{L-1}-\frac{r_{\min}(\delta')}{L-1} + \frac{L}{L-1}(\delta'-\delta)=G(\delta').
    \end{equation}

    This is the desired result.
    
    Conversely, if $f$ is $\frac{L}{L-1}$-weakly convex, since $\frac{L}{L-1} >1$ and 
    \begin{equation}
        \frac{\frac{L}{L-1}}{\frac{L}{L-1}-1}=L,
    \end{equation}
    
    \Cref{lem:lipschitz_from_weak_convex} yields that 
    \begin{equation}
        r_{\min} \text{ is $L$-Lipschitz continuous}.
    \end{equation}

    Now for the second equivalence, if $r_{\min}$ is $1$-Lipschitz continuous, it is $L$-Lipschitz continuous for any $L>1$. 
    
    Hence, 
    \begin{equation}
        r\mapsto f(r)+\left(1-\frac{1}{L}\right)\frac{r^2}{2} \text{ is convex for any } L>1.
    \end{equation}
    
    This family of convex functions converges pointwise to $f$ as $L\rightarrow 1^+$, which is thus a convex function.

    Conversely, if $f$ is convex, it is $V_0$-weakly-convex for any $V_0>1$. Therefore, 
    
    \begin{equation}
        r_{\min}\text{ is } \frac{V_0}{V_0-1} \text{-Lipschitz continuous}.
    \end{equation} 
    
    Taking the point wise limit $V_0\rightarrow \infty$ yields that $r_{\min}$ is $1$-Lipschitz continuous.
\end{proof}

\begin{corollary}\label{cor:preimage_characterization}
Suppose $E_f = \mathbb{R}$ and $f$ is strictly $1$-weakly convex. Let $r_{\min}$ be the unique continuous proximal selection of $f$. For $r_0 \in \mathbb{R}$,
$$r_{\min}^{-1}(r_0) = r_0 + [f'_-(r_0), f'_+(r_0)].$$
\end{corollary}

\begin{proof}[\Cref{cor:preimage_characterization}]
By Corollary~\ref{cor:continuity_equiv}, $r_{\min}$ is uniquely determined and onto. By Proposition~\ref{prop:si_characterization},
$$r_{\min}^{-1}(r_0) = \{\delta \in \mathbb{R} : r_0 \in \mathcal{P}_f(\delta)\} = \{\delta \in \mathbb{R} : s_f(r_0) \leq \delta \leq i_f(r_0)\} = [s_f(r_0), i_f(r_0)].$$

By Proposition~\ref{prop:weak_convex_si}, $s_f(r_0) = r_0 + f'_-(r_0)$ and $i_f(r_0) = r_0 + f'_+(r_0)$.
\end{proof}

\vspace{1em}
We are now ready to state the proof of the main results of this appendix.
\begin{proof}[\Cref{thm:main_proximal_characterization}]
The proof follows by combining the previous results:

The first statement follows from Proposition~\ref{prop:existence_proximal}. The non-decreasing property and almost everywhere differentiability follow from Proposition~\ref{prop:convexity_moreau}.

The three main equivalences follow from, Corollary~\ref{cor:continuity_equiv} for the continuity equivalence and Corollary~\ref{cor:lipschitz_equiv} for both Lipschitz equivalences.

The uniqueness statement follows from any of these conditions implying strict $1$-weak convexity (or stronger), which by Corollary~\ref{cor:strict_weak_convex} and Proposition~\ref{prop:si_characterization} implies $\mathcal{P}_f$ is singleton-valued.

The characterization of $r_{\min}^{-1}(r_0)$ follows from Corollary~\ref{cor:preimage_characterization}.

Finally, $r_{\min}$ is increasing if and only if it is injective, which occurs if and only if $r_{\min}^{-1}(r_0)$ is a singleton for all $r_0$. This happens if and only if $f'_-(r_0) = f'_+(r_0)$ for all $r_0$, i.e., $f$ is differentiable everywhere.
\end{proof}

\vspace{1em}
\begin{proof}[\Cref{cor:scaled_proximal}]
The proof of the first statement follows by applying ~\Cref{lem:lipschitz_from_weak_convex} to $Vf$ since $V_0/V>1$. The second statement follows from ~\Cref{thm:main_proximal_characterization} applied to the function $Vf$ in place of $f$, with $L= \frac{V_0}{V_0-V}>1$.

Under~\Cref{ass:lossregfun}, for the loss $\mathscr{L}$, the function $z \mapsto \mathscr{L}(y,z)$ is $V_L$-weakly convex. 

Under~\Cref{ass:weak_convexity}, applying~\Cref{cor:scaled_proximal} with $V = \frac{\tau^\star}{\kappa^\star}< V_0 \defeq V_L$ and $L = L_L$ guarantees the existence of a unique $L_L$-Lipschitz continuous selection $\tilde r_L$ such that

$$\forall \delta\in \RR, \: \tilde{r}_L(\delta) \in \argmin_{r \in \mathbb{R}} \left\{\frac{(\delta-r)^2}{2} + \frac{\tau^\star}{\kappa^\star}\lossfun(y,r)\right\}.$$

If we define 
$\mathcal{P}_ {\frac{\tau^\star}{\kappa^\star} \mathscr{L}(y,\cdot)} : \omega \in \mathbb{R} \mapsto \tilde r_L(\omega)\in\mathbb{R}$ satisfying $\mathcal{P}_{\frac{\tau^\star}{\kappa^\star} \lossfun(y,\cdot)}(\omega) \in \mathcal{P}_{\frac{\tau^\star}{\kappa^\star}\mathscr{L}(y,\cdot)}(\omega)$ for all $(\omega,y) \in \mathbb{R}^2$. 

Similarly, for the regularization $\mathscr{R}$, the function $w \mapsto \mathscr{R}(w)$ is $V_R$-weakly convex. 

\Cref{cor:scaled_proximal} with $V = \frac{1}{\eta^\star}$ and $L = L_R$ provides a unique $L_R$-Lipschitz continuous selection $\mathcal{P}_{\frac{1}{\eta^\star}\mathscr{R}(\cdot)} : \mathbb{R} \to \mathbb{R}$ with $\mathcal{P}_{\frac{1}{\eta^\star}\mathscr{R}(\cdot)}(w) \in \mathcal{P}_{\frac{1}{\eta^\star}\mathscr{R}(\cdot)}(w)$ for all $w \in \mathbb{R}$. These Lipschitz-continuous proximal selections define the update functions $g^\star$ and $f^\star$ in~\cref{eq:output-input-denoisers-amp}, recovering the regularity conditions originally stated in the first formulation of our framework.
\end{proof}

\section{Applications}\label{sec:app:proofs-applications}
\subsection{Examples of Weakly Convex Losses}

We verify \Cref{ass:weak_convexity} for standard robust losses found in the statistics literature.

\begin{example}[Huber Loss]
The Huber loss is defined as
$$\mathscr{L}_H^\xi(y,z) = \xi^2 \int_0^{|y-z|/\xi} \min(1,x)\,dx.$$
Its second derivative with respect to $z$ is
$$\frac{\partial^2 \mathscr{L}_H^\xi}{\partial z^2}(y,z) = \begin{cases} 1 & \text{if } |y-z| \leq \xi \\ 0 & \text{if } |y-z| > \xi \end{cases} \geq 0.$$
Thus $z \mapsto \mathscr{L}_H^\xi(y,z)$ is convex, hence $V_0$-weakly convex for any $V_0 \geq 0$.
\end{example}

\begin{example}[Tukey Biweight Loss]
The Tukey biweight loss is defined as
$$\mathscr{L}_T^\xi(y,z) = \begin{cases}
\frac{\xi^2}{6}\left[1 - \left(1 - \frac{(y-z)^2}{\xi^2}\right)^3\right] & \text{if } |y-z| \leq \xi \\
\frac{\xi^2}{6} + \varrho|y-z - \xi|^3 & \text{if } y-z > \xi \\
\frac{\xi^2}{6} + \varrho|y-z + \xi|^3 & \text{if } y-z < -\xi
\end{cases}$$
where $\varrho > 0$ is a coercivity parameter. For $u := y-z$ with $|u| \leq \xi$, the second derivative is
$$\frac{\partial^2 \mathscr{L}_T^\xi}{\partial z^2}(y,z) = \left(1 - \frac{u^2}{\xi^2}\right)\left(1 - 5\frac{u^2}{\xi^2}\right).$$
This is negative when $\frac{u^2}{\xi^2} \in (\frac{1}{5}, 1)$. To find the minimum, set $t = \frac{u^2}{\xi^2}$ and minimize $(1-t)(1-5t) = 1 - 6t + 5t^2$. Taking the derivative yields $-6 + 10t = 0$, giving $t = \frac{3}{5}$. At this point, the second derivative equals $(1 - \frac{3}{5})(1 - 3) = -\frac{4}{5}$. For $|u| > \xi$, we have $\frac{\partial^2 \mathscr{L}_T^\xi}{\partial z^2} = 6\varrho|u - \xi| > 0$.

Taking $V_0 = \frac{5}{4}$ ensures 
$$\frac{\partial^2}{\partial z^2}\left(V_0 \mathscr{L}_T^\xi + \frac{z^2}{2}\right) \geq \frac{5}{4} \cdot \left(-\frac{4}{5}\right) + 1 = 0.$$
Thus $\mathscr{L}_T^\xi$ is $\frac{5}{4}$-weakly convex.
\end{example}

\begin{example}[Cauchy Loss]
The Cauchy loss is defined as
$$\mathscr{L}_C^\xi(y,z) = \frac{\xi^2}{2}\log\left(1 + \frac{(y-z)^2}{\xi^2}\right).$$
Its second derivative with respect to $z$ is
$$\frac{\partial^2 \mathscr{L}_C^\xi}{\partial z^2}(y,z) = \frac{1 - \frac{(y-z)^2}{\xi^2}}{\left(1 + \frac{(y-z)^2}{\xi^2}\right)^2}.$$
Setting $t = \frac{(y-z)^2}{\xi^2}$ and minimizing $h(t) = \frac{1-t}{(1+t)^2}$ gives $h'(t) = \frac{t-3}{(1+t)^3} = 0$ at $t = 3$, with minimum value $h(3) = \frac{-2}{16} = -\frac{1}{8}$.

Taking $V_0 = 8$ ensures
$$\frac{\partial^2}{\partial z^2}\left(V_0 \mathscr{L}_C^\xi + \frac{z^2}{2}\right) \geq 8 \cdot \left(-\frac{1}{8}\right) + 1 = 0,$$
so $\mathscr{L}_C^\xi$ is $8$-weakly convex.
\end{example}

\begin{example}[$\ell_2$ Regularization]
The $\ell_2$ regularization is defined as
$\mathscr{R}_\lambda(w) = \frac{\lambda}{2}w^2$.
Its second derivative is
$\mathscr{R}_\lambda''(w) = \lambda \geq 0$
for $\lambda > 0$. This is convex, hence $V_0$-weakly convex for all $V_0 \geq 0$.
\end{example}

\subsection{Negative Regularization in a Spherical Shell}

To study the landscape of the problem at hand we slice it in spherical shells of radius squared equal to $q$. We have that
\begin{align}
    \min_{\w \in \RR^d} \frac{1}{d} \qty[ 
        \sum_{\dataidx=1}^n 
        \lossfun\qty(y_\dataidx, \frac{\w^\top \x_\dataidx}{\sqrt{d}} )  
    ] + \frac{\lambda}{2} \frac{\norm{\w}_2^2}{d} 
    &= \min_{q \geq 0} \min_{\w \in \mathcal{K}_{q,q}} \frac{1}{d} \qty[ 
        \sum_{\dataidx=1}^n 
        \lossfun\qty(y_\dataidx, \frac{\w^\top \x_\dataidx}{\sqrt{d}} )  
    ] + \frac{\lambda}{2} q \\
    &=: \min_{q \geq 0} h(q) + \frac{\lambda}{2} q 
\end{align}
where the second minimization can be studied with the help of~\Cref{thm:main}. We now are left with the study of a function in a single variable $h(q)$.

We consider the specific form of the optimization objective obtained from~\Cref{thm:moreau-form-energy}, where for $(m^\star, q^\star, \tau^\star, \kappa^\star, \nu^\star, \eta^\star) \in \mathcal{S}^\star$ we have that
\begin{equation}\label{eq:simpl-opt-value-linear-loss}
    h(q) = \alpha \mathbb{E}\qty[
        \lossfun\qty(y, \mathcal{P}_{\frac{\tau^\star}{\kappa^\star}\lossfun(y,\cdot)} \qty(\frac{m^\star}{\sqrt{\rho}} s + \sqrt{q  - \frac{(m^\star)^2}{\rho}} h))
    ] \,,
\end{equation}
and where $q^\star = q$ since in the problem definition we have that $a = b = q$. We also note that because of the assumption on $\lossfun$ $h(q)$ is also convex.

We study first the derivative of the first term. We notice three facts: first, the value of the derivative in zero is non-positive, secondly it is bounded from above by a constant and third for $q\to\infty$ the derivative reaches zero from above.

We thus define the value of $\lambda_c$ as
\begin{equation}
    \lambda_c = \min \qty{\lambda \in \RR \mid \exists q \geq 0 \quad \partial h(q) = \frac{\lambda}{2} } \,, 
\end{equation}

The presence of a single minimizer for $\lambda \geq 0$ follows from the convexity of the optimization objective.
For $\lambda_c \leq \lambda <0$ follows from the fact that for $\lambda \in [\lambda_c, 0]$ there are two values of $q$ such that $\partial h(q) = \frac{\lambda}{2}$.
If we consider $\lambda < \lambda_c$ we have that there is no stationary point for $h(q)$ and thus it is a decreasing function of $q$ and its minimum is attained at $q \to \infty$.

\subsection{Performances of Linearized AMP}

The equations presented in the main text are the ones presented in~\Cref{cor:extr_se} where 
the projection on the set $\mathcal{K}_{1,1}$ at each step simplifies the second and third line in~\cref{eq:app:self-consistent-eqs} to be
\begin{equation}
    m = \frac{\hat{m}}{\hat{q} + \hat{m}^2} \,, \quad \tau = \frac{1}{\sqrt{\hat{q} + \hat{m}^2}} \,.
\end{equation}

\section{Details on Numerical Experiments}\label{sec:app:numerical-experiments}

To find the minimum of the~\cref{eq:min-risk-limit-def} we use the fixed point characterization from~\Cref{cor:extr_se} as they are written in a way amenable to be solved via fixed-point
iteration. Starting from a random initialization, we iterate through both the hat and non-hat variable equations until the maximum absolute difference between the order parameters in two successive iterations falls below a tolerance of
$10^{-5}$.

To speed-up convergence we use a damping scheme, updating each order parameter at iteration $i$, designated as
$x_i$, using $x_i := x_i \mu + x_{i-1}(1 - \mu)$, with $\mu$ as the damping parameter.

Once convergence is achieved for fixed value of hyper-parameters they are optimized using a gradient-free numerical minimization procedure for a one dimensional minimization.

For each iteration, we evaluate the proximal operator numerically using SciPy's Brent's algorithm for root finding (\texttt{scipy.optimize.minimize\_scalar}).
The numerical integration is handled with SciPy's quad method (\texttt{scipy.integrate.quad}), which provides adaptive quadrature of a given function over a specified interval. These numerical techniques allow us to evaluate the equations and perform the necessary integrations with the desired accuracy. 

Concerning the setting of each figure shown in the main text we have
\begin{description}
\item[\Cref{fig:tukey-huber-comparison}] The left figure is a plot of the different loss functions defined in~\Cref{sec:non-convex-losses} with $\xi = 1.5$. 
The data model considered in the right figure is
\begin{equation}\label{eq:app:data-mod}
    \pout{y \mid z} = \begin{cases}
        z + \sqrt{\Delta} z_1 & \text{with prob. } 1-\epsilon \\
        \sqrt{\Sigma} z_2 & \text{with prob. } \epsilon 
    \end{cases}
\end{equation}
with $\epsilon = 0.1$, $\Delta = \Sigma = 1$, and $z_1,z_2 \sim \mathcal{N}(0,1)$ independent. 
The error bars are 10 different realizations of the numerical minimization of the ERM problem with $d=1000$.
\item[\Cref{fig:negative-regularisation}] The figure on the left shows the minimum value of $\minriskfun$ as a function of the fixed $q$. The channel chosen is the one in~\cref{eq:app:data-mod} with $\epsilon = 0.3$, $\Delta = 1.0$, $\Sigma = 5.0$. We use the Huber loss with $\xi = 1$. The values of the regularization parameter chosen are $\lambda \in \qty{-1.0, 1.0, -2.1, -1.735, 0.0}$ and $\alpha = 10$.
For the right figure we use again the channel defined in~\cref{eq:app:data-mod} with $\lambda = $ and the previous parameter as before. The gradient descent iterations are done on a system with $d = 1000$.
\end{description} 

\newpage

\subsection*{Acknowledgements}
MV would like to thank Emanuele Troiani and Vittorio Erba for discussion at the beginning of the project. 
We thank also Vanessa Piccolo for careful rereading of the manuscript.
This work was supported by the Swiss National Science Foundation under grant SNSF OperaGOST (grant number 200390).

\bibliographystyle{unsrtnat}
\bibliography{refs}

\end{document}

%% file: math-definitions.tex
\DeclareMathOperator*{\argmin}{arg\,min}

\newcommand{\outputfun}{\ensuremath{\varphi}}
\newcommand{\pout}[1]{\ensuremath{P_{\mathrm{out}}\qty(#1)}}

\newcommand{\defeq}{\mathrel{\mathop:}=}

\newcommand{\what}{\ensuremath{\hat{\mathbf{w}}}}
\newcommand{\w}{{\ensuremath{\mathbf{w}}}}

\newcommand{\x}{\ensuremath{\mathbf{x}}}
\newcommand{\y}{\ensuremath{\mathbf{y}}}
\newcommand{\z}{{\ensuremath{\mathbf{z}}}}

\newcommand{\p}{{\ensuremath{\mathbf{p}}}}

\newcommand{\s}{{\ensuremath{\mathbf{s}}}}

\newcommand{\vv}{\ensuremath{\mathbf{v}}}

\newcommand{\f}{\ensuremath{\mathbf{f}}}

\newcommand{\xib}{\ensuremath{\boldsymbol{\xi}}}

\newcommand{\dataset}{\ensuremath{\mathcal{D}}}

\newcommand{\gaussvecone}{\ensuremath{\mathbf{g}}}
\newcommand{\gaussvectwo}{\ensuremath{\mathbf{h}}}

\newcommand{\datamat}{\ensuremath{\boldsymbol{X}}}

\newcommand{\dataidx}{\ensuremath{\mu}}
\newcommand{\dimidx}{\ensuremath{i}}

\newcommand{\riskfun}{\ensuremath{\mathcal{A}}}
\newcommand{\energyfun}{\ensuremath{\mathcal{E}}}
\newcommand{\minriskfun}{\ensuremath{{\riskfun}^\star}}
\newcommand{\wminimizer}{\ensuremath{\hat{\w}}}
\newcommand{\wteacher}{\ensuremath{\tilde{\w}}}
\newcommand{\minauxprob}{\ensuremath{\tilde{A}}}

\newcommand{\lossfun}{\ensuremath{\mathscr{L}}}
\newcommand{\regfun}{\ensuremath{\mathscr{R}}}

\newcommand{\outden}[1][\ ]{{\ensuremath{g^{#1}}}}
\newcommand{\inden}[1][\ ]{{\ensuremath{f^{#1}}}}

\newcommand{\EE}[1]{\ensuremath{\mathbb{E}\qty[#1]}}

\newcommand{\RR}{\ensuremath{\mathbb{R}}}
\newcommand{\N}{\ensuremath{\mathbb{N}}}
